\documentclass[%
  aip,
  cha,
  reprint,
  amsmath,
  amssymb,
  superscriptaddress,
  floatfix,
]{revtex4-2}

\usepackage{graphicx}%
\usepackage{dcolumn}%
\usepackage{bm}%

\usepackage[utf8]{inputenc}
\usepackage[T1]{fontenc}
\usepackage{etoolbox}

\usepackage{hyperref}
\usepackage{url}
\usepackage[capitalize]{cleveref}
\usepackage{booktabs}
\usepackage{multirow}
\usepackage{amsfonts}
\usepackage{amsthm}
\usepackage{nicefrac}
\usepackage{microtype}
\usepackage{xcolor}
\usepackage{tikz}
\usetikzlibrary{decorations.markings,decorations.pathmorphing,calc,3d,arrows.meta}
\usepackage{siunitx}
\usepackage{mathtools}

\newtheorem{proposition}{Proposition}

\DeclareMathOperator{\Real}{Re}
\DeclareMathOperator{\Imag}{Im}
\DeclareMathOperator*{\argmin}{arg\,min}

\newcommand{\sR}{\mathbb{R}}

\begin{document}

\title[Projected Neural Differential Equations]{Projected Neural Differential Equations for Learning Constrained Dynamics}

\author{Alistair White}
\email{alistair.white@tum.de}
\affiliation{Munich Climate Center and Earth System Modelling Group, Department of Aerospace and Geodesy, School of Engineering and Design, Technical University of Munich, Munich, Germany}
\affiliation{Potsdam Institute for Climate Impact Research, Potsdam, Germany}

\author{Anna Büttner}
\affiliation{Potsdam Institute for Climate Impact Research, Potsdam, Germany}

\author{Maximilian Gelbrecht}
\affiliation{Potsdam Institute for Climate Impact Research, Potsdam, Germany}
\affiliation{Munich Climate Center and Earth System Modelling Group, Department of Aerospace and Geodesy, School of Engineering and Design, Technical University of Munich, Munich, Germany}

\author{Valentin Duruisseaux}
\affiliation{California Institute of Technology, Pasadena, California, USA}

\author{Niki Kilbertus}
\affiliation{Technical University of Munich, Munich, Germany}
\affiliation{Helmholtz Munich, Neuherberg, Germany}
\affiliation{Munich Center for Machine Learning (MCML), Munich, Germany}

\author{Frank Hellmann}
\affiliation{Potsdam Institute for Climate Impact Research, Potsdam, Germany}

\author{Niklas Boers}
\affiliation{Munich Climate Center and Earth System Modelling Group, Department of Aerospace and Geodesy, School of Engineering and Design, Technical University of Munich, Munich, Germany}
\affiliation{Potsdam Institute for Climate Impact Research, Potsdam, Germany}
\affiliation{University of Exeter, Exeter, United Kingdom}

\date{\today}

\begin{abstract}
  Neural differential equations offer a powerful approach for learning dynamical systems from data.
  However, they do not inherently respect known constraints, such as conservation laws, that should be obeyed by the learned dynamics.
  It is well known that enforcing constraints in data-driven models can enhance their generalizability and numerical stability.
  In this paper, we introduce \emph{projected neural differential equations} (PNDEs), a method for constraining neural differential equations based on projection of the predicted velocities onto the tangent space of the manifold that fulfills the constraint.
  In tests on several examples from different fields, including chaotic dynamical systems and power grid models, PNDEs outperform existing methods for constraining learned dynamics, require fewer hyperparameters, and are computationally more efficient.
  The proposed approach demonstrates potential for enhancing the modeling of constrained dynamical systems, particularly in domains where accuracy and reliability are essential.
\end{abstract}

\maketitle

\begin{quotation}
  Machine learning models that simulate physical systems often struggle to respect fundamental constraints, such as mass, momentum, or energy conservation.
  This limitation can lead to unphysical, inaccurate, and unstable predictions.
  We propose projected neural differential equations (PNDEs), a method that mathematically guarantees learned models will satisfy known physical constraints by projecting their predictions onto the space of physically valid solutions.
  Tests on challenging systems—including highly chaotic dynamical systems and complex power grid models—demonstrate that PNDEs enforce constraints more accurately than existing approaches while being more efficient and requiring fewer hyperparameters.
  By enabling machine learning models to reliably incorporate domain knowledge, PNDEs offer a promising path toward accurate and efficient simulation of complex physical systems in applications ranging from grid stability analysis to climate modeling, particularly in safety-critical contexts where constraint violations are unacceptable.
\end{quotation}

\section{Introduction}
\label{sec:intro}
Numerical simulation of dynamical systems plays a pivotal role in science and engineering.
Across domains, simulations extend the reach of scientific inquiry far beyond the limitations of direct observation and experimentation.
However, as the scale and complexity of the systems being studied increases, the computational cost of traditional numerical methods can become prohibitive.
Deep learning surrogates, enabled by the emergence of efficient architectures and specialized accelerator hardware, offer promising alternatives.
By circumventing the computational limitations of conventional approaches, these methods stand to further advance the use of simulations for scientific inquiry.\cite{kochkov_machine_2021,jumper_highly_2021,pathak_fourcastnet_2022,kovachki_neural_2023,merchant_scaling_2023,azizzadenesheli_neural_2024}

\begin{figure}[t]
  \centering
\begin{tikzpicture}[
        x={(0.86625cm,-0.50025cm)},
        y={(0.86625cm,0.50025cm)},
        z={(0cm,0.99975cm)},
        >=Stealth
    ]
    \draw[thick,fill=blue!15,opacity=0.8]
    (0,0,0) to[out=60,in=180] (2,1,1)
    to[out=0,in=120] (4,0,0.5)
    to[out=300,in=0] (2,-1,-0.5)
    to[out=180,in=240] cycle;
    \shade[ball color=blue!5,opacity=0.2]
    (0,0,0) to[out=60,in=180] (2,1,1)
    to[out=0,in=120] (4,0,0.5)
    to[out=300,in=0] (2,-1,-0.5)
    to[out=180,in=240] cycle;
    \coordinate (u) at (2.5,0.5,0.75);
    \filldraw[fill=red!20,opacity=0.7,draw=red!50!black]
    ($(u)+(-1.5,-1.5,0)$) -- ($(u)+(1.5,-1.5,0)$) --
    ($(u)+(1.5,1.5,0)$) -- ($(u)+(-1.5,1.5,0)$) -- cycle;
    \coordinate (F) at ($(u)+(0.8,0.6,0.7)$);
    \draw[->,line width=1.35pt,red!80!black] (u) -- (F) ; %
    \node[red!80!black,font=\small] at ($(u)+(0.4,0.2,0.65)$) {$\bar f_{\theta}(u)$};
    \coordinate (F_parallel) at ($(u)+(0.8,0.6,0)$);
    \coordinate (F_perp) at ($(F)-(F_parallel)$);
    \draw[orange,line width=1.35pt,
        decoration={markings,
                mark=between positions 0.25 and 0.99 step 0.2 with {\arrow[scale=0.9]{>}}},
        postaction={decorate}]
    (0.7,0.15,0.3) .. controls (1.6,0.7,0.85) and (2.1,0.4,0.7) ..
    (u) .. controls ($(u)!0.5!(F_parallel)$) and (3.1,0.6,0.8) ..
    (3.6,0.1,0.4) .. controls (3.8,-0.1,0.3) .. (3.9,-0.2,0.25);
    \draw[->,line width=1.35pt,green!50!black] (u) -- (F_parallel) ; %
    \node[green!50!black,font=\small] at ($(u)+(1.5,0.2,0.3)$) {$\mathrm{Proj}_u \bar f_{\theta}(u)$};
    \draw[->,dashed,gray,semithick] (F) -- (F_parallel) ; %
    \node[font=\large] at (2,-1.7,-0.3) {$\mathcal M$};
    \node[font=\large] at (4.95,1.35,1.1) {$\text{T}_u \mathcal M$};
    \fill[black] (u) circle (2.25pt);
    \node[below left,font=\normalsize] at (u) {$u$};
    \coordinate (u0) at (0.7,0.15,0.3);
    \fill[orange] (u0) circle (2.25pt);
    \node[below left,font=\normalsize] at (u0) {$u_0$};
\end{tikzpicture}   \caption{Schematic of projected neural differential equations (PNDEs).
    The vector field of the unconstrained NDE (red arrow) is projected to the tangent space $\text{T}_u\mathcal{M}$ of the constraint manifold $\mathcal{M}$.
    For initial conditions $u_0 \in \mathcal{M}$, solutions (orange) of the projected vector field (green arrow) remain on the manifold and thereby satisfy the constraints.
  }
  \label{fig:nde}
\end{figure}

Among existing machine learning approaches, neural ordinary differential equations (NODEs) offer a promising approach for data-driven modeling of dynamical systems, combining the expressiveness of neural networks with the interpretability of differential equations.\cite{chen_neural_2018}
Building upon NODEs, universal differential equations (UDEs) offer a hybrid framework that integrates neural networks with differentiable physical models.\cite{rackauckas_universal_2021}
We refer to these approaches collectively as~\emph{neural differential equations} (NDEs).
NDEs have demonstrated significant potential across various scientific disciplines, including weather and climate modeling,\cite{kochkov_neural_2024,yu_climsim-online_2024,gelbrecht_differentiable_2023,gelbrecht_pseudospectralnet_2025} robotics,\cite{richards_adaptive-control-oriented_2021} epidemiology,\cite{dandekar_machine_2020} bioengineering,\cite{narayanan_hybrid_2022} and chaotic systems.\cite{schotz_machine_2026}
Despite excellent performance within the training data, NDEs often generalize poorly, for example, to new initial conditions or future times.\cite{linot_stabilized_2023,jia_feedback_2025,yin_leads_2021}
Due to the iterative nature in which neural networks are applied within NDEs, errors quickly accumulate and can push the trajectory far outside the training distribution, where performance degrades.

Inductive biases play an important role in avoiding such issues.
In scientific machine learning, for example, inductive biases can take the form of physical constraints, such as conservation laws or other domain-specific knowledge, that can enhance model performance and interpretability.\cite{Karniadakis2021}
Physical constraints are often incorporated as additional loss terms, for example, in physics-informed neural networks\cite{Raissi2017Part1,Raissi2017Part2,raissi_physics-informed_2019} and physics-informed operator learning.\cite{PINO,WangPIDeepONet,Goswami2023,lin2025mGNO,ganeshram2025fcpinohighprecisionphysicsinformed}
However, a variety of issues can arise with these \textit{soft} constraints.
For instance, the resulting composite loss can be expensive to evaluate, due to derivatives in the physics loss, and challenging to minimize.
Furthermore, a number of numerical issues are known to arise, such as worse conditioning, conflicting loss terms, inaccuracies in the approximated derivatives, and the existence of spurious solutions satisfying the constraints on the collocation points of the physics loss.\cite{Leiteritz2021,Krishnapriyan2021,huang2023limitations,Wang2021,Wang2022_3,Rohrhofer2022}
Yet perhaps the most significant drawback of soft constraints is that they provide no guarantees that the physical constraints will be satisfied at inference time; at best, they can only hope to guide the model weights toward physically plausible solutions during training.
This is inadequate for safety-critical applications---such as power grid modeling---where provable, robust adherence to physical constraints is essential.
Furthermore, long-horizon simulations often demand strict constraint satisfaction: in climate modeling, for example, small per-timestep energy conservation errors can accumulate over time and introduce significant biases in the simulated long-term climate.
These considerations motivate stricter enforcement of constraints, both during training and at inference time, in the form of \emph{hard} constraints.
\paragraph{Contribution.}
In this paper, we introduce \emph{projected neural differential equations} (PNDEs), a method for incorporating hard constraints in NDEs based on projection of the NDE's predicted velocities onto tangent spaces of the constraint manifold (see Fig.~\ref{fig:nde}).
Since the constraint manifold is defined by algebraic constraints on the system state, our method can enforce diverse constraints, such as conservation laws and holonomic constraints, without requiring the system to be measured in specific (e.g., generalized or canonical) coordinates.
To enable efficient training of PNDEs, we derive analytic expressions for the gradients of the projection operator and provide efficient batched implementations for both forward and backward passes.
Our method enforces constraints to near machine precision---several orders of magnitude more accurately than existing methods---while requiring fewer hyperparameters and avoiding the numerical stiffness that can affect other approaches.
We further demonstrate that explicit constraints enable learning of complex dynamics in simpler coordinate systems (e.g., Cartesian) that would otherwise be intractable in their natural, generalized coordinates.
\paragraph{Organization.}
The remainder of this paper is organized as follows.
Section~\ref{sec:pnde} derives PNDEs and presents efficient implementations.
Section~\ref{sec:related_work} positions our approach within the broader literature on constrained learning.
Section~\ref{sec:experiments} validates the method on chaotic dynamical systems and power grid models, demonstrating superior constraint satisfaction and numerical stability.

\section{Projected Neural Differential Equations}
\label{sec:pnde}

\subsection{Background}
The key geometric insight of our approach is that constraining solutions to a manifold $\mathcal{M}$ requires constructing a vector field tangent to $\mathcal{M}$ everywhere.
We briefly review the relevant differential geometry, referring readers to existing literature for a more detailed treatment.\cite{Lang1999,lee_introduction_2012,boumal_introduction_2024}

The set of tangent vectors to a manifold $\mathcal{M}$ at a point $u\in \mathcal{M}$ is a vector space called the \emph{tangent space} to $\mathcal{M}$ at $u$ and is denoted by $\text{T}_u \mathcal{M}$.
The disjoint union of all tangent spaces to $\mathcal{M}$ forms the \emph{tangent bundle} $\text{T} \mathcal{M} = \{  (u,v) \ | \  u\in \mathcal{M},  \ v \in \text{T}_u \mathcal{M}  \}$ of $\mathcal{M}$.
A \emph{vector field} on a manifold~$\mathcal{M}$ is a continuous map~$X:\mathcal{M} \rightarrow \text{T} \mathcal{M}$ with the property that $X(u) \in \text{T}_u \mathcal{M}$ for all~$u \in \mathcal{M}$.
The \emph{integral curve} at $u$ of a vector field~$X$ on~$\mathcal{M}$ is the smooth curve~$\gamma$ on~$\mathcal{M}$ such that $\gamma(0)=u$ and $\gamma'(t) = X(\gamma(t))$ for all~$t$ in the interval of interest.
In other words, the integral curve is the solution of the ODE $\gamma'(t) = X(\gamma(t))$ with $\gamma(0) = u$, so theorems for the existence, uniqueness, and smoothness of ODE solutions translate directly to integral curves of vector fields on manifolds.

Crucially, if $\mathcal{M}$ is embedded in an ambient space and $X(u)$ lies in the tangent space $\text{T}_u \mathcal{M}$ for all $u$, then the integral curve starting at any $u_0 \in \mathcal{M}$ remains on $\mathcal{M}$ for all time.
This forms the foundation of our projection-based approach.
\subsection{Problem Setting}
\paragraph{Neural Differential Equations.}
Assume we have an ambient $n$-dimensional Euclidean phase space~$\mathcal E$, and consider the neural differential equation
\begin{equation}
  \label{eq:nde}
  \dot u = f_{\theta}(u(t),t) \qquad u(0) = u_0,
\end{equation}
where $u: \sR \to \mathcal E$, $u_0\in \mathcal{E}$ is an initial condition, $\dot{u}=du/dt$, and $f_{\theta}$ is parameterized by a neural network with parameters~$\theta \in \sR^d$.

Given observations $\{t_i, u(t_i)\}_{i=1}^N$, the parameters~$\theta$ can be optimized using stochastic gradient descent by integrating a trajectory $\hat u(t) = \mathrm{ODESolve}(u_0, f_{\theta}, t)$ and taking gradients of the loss~$\mathcal{L}(u, \hat u)$ with respect to~$\theta$.
Gradients of the $\mathrm{ODESolve}$ operation can be obtained using adjoint sensitivity analysis (sometimes called \emph{optimize-then-discretize}) or via direct automatic differentiation of the solver operations (\emph{discretize-then-optimize}).\cite{chen_neural_2018,kidger_neural_2022}

Since any non-autonomous system can be equivalently represented as an autonomous system by adding time as an additional coordinate, for ease of notation we will drop the time-dependence in Eq.~(\ref{eq:nde}) and consider only autonomous systems from now on, without loss of generality.

\paragraph{Constraints.}
Suppose that the state $u(t)$ of the system must be constrained to the set
\begin{equation}
  \label{manifold}
  \mathcal M = \{u \in \mathcal E \,;\, g(u) = 0 \},
\end{equation}
where $g \colon \mathcal E \to \sR^m$ with $g : u \mapsto \big(g_1(u), \ldots, g_m(u)\big)$, given $m<n$ independent explicit algebraic constraints~$g_i(u) =0$.
Assuming that the Jacobian~$\mathrm{D}g(u)$ exists and has full rank for all~$u \in \mathcal M$, the Regular Level Set Theorem\cite{lee_introduction_2012} implies that $\mathcal M$ is a smooth $(n-m)$-dimensional embedded submanifold of~$\sR^n$.
To satisfy the constraints $g(u) = 0$, we seek solutions of Eq.~(\ref{eq:nde}) that are constrained to the embedded submanifold~$\mathcal M$, which we refer to as the \emph{constraint manifold}.
In other words, we seek integral curves on~$\mathcal{M}$.

\subsection{Proposed Approach}

We want to learn $f_{\theta} : \mathcal{M} \to \mathrm{T}\mathcal{M}$ as a smooth vector field on the embedded submanifold~$\mathcal{M}$ with local defining function $g$.
We realize this by first parameterizing a smooth vector field~$\bar f_{\theta} \colon \mathcal{E} \to \mathcal{E}$ on the ambient Euclidean space using a neural network, and then projecting~$\bar f_{\theta}(u)$ using a suitable projection operator~$\mathrm{Proj}_u \colon \mathcal{E} \to \mathrm{T}_u\mathcal{M}$ to obtain the desired tangent vector,
\begin{equation}
  \label{eq:projected_field}
  f_{\theta}(u) = \mathrm{Proj}_u\bigl(\bar f_{\theta}(u)\bigr) \in \mathrm{T}_u\mathcal{M}.
\end{equation}
Since $\mathcal M$ is a smooth embedded submanifold of~$\mathcal E$, the restriction~$\bar f_\theta \vert_{\mathcal{M}}$ is also smooth.
Note that the projection acts fiberwise (i.e., acting ``point-by-point''); for each~$u \in \mathcal{M}$, it projects the vector~$\bar f_{\theta}(u)$ at~$u$ onto~$\mathrm{T}_u\mathcal{M}$.

\paragraph{Construction of the projection operator.}
We now derive a suitable projection operator, following the approach of \citet{boumal_introduction_2024}.
Given the standard Euclidean inner product~$\langle u, v \rangle = u^\top v$ and its induced norm~$\lVert \cdot \rVert_2$ on~$\mathcal{E} = \mathbb{R}^n$, the differential~$\mathrm{D}g(u)$ and its adjoint~$\mathrm{D}g(u)^*$ define linear maps
\begin{subequations}
  \begin{align}
    & \mathrm{D}g(u)[v] = \bigl( \langle \nabla g_1(u), v \rangle, \ldots, \langle \nabla g_m(u), v \rangle \bigr), \\
    & \mathrm{D}g(u)^*[\alpha] = \sum_{i=1}^{m} \alpha_i \nabla g_i(u).
  \end{align}
\end{subequations}
Denoting the orthogonal projection from~$\mathcal{E}$ to~$\mathrm{T}_u \mathcal{M}$ by~$\mathrm{Proj}_u : \mathcal{E} \to \mathrm{T}_u \mathcal{M}$, we can uniquely decompose any vector~$v \in \mathcal{E}$ into its components parallel and perpendicular to~$\mathrm{T}_u \mathcal{M}$,
\begin{equation}
  \label{eq:v_decompose}
  v = \mathrm{Proj}_u(v) + \mathrm{D}g(u)^*[\alpha].
\end{equation}
The coefficients~$\alpha$ are obtained as the solution of the least-squares problem
\begin{equation}
  \label{eq:alpha}
  \alpha = \argmin_{\alpha \in \sR^m} \lVert v - \mathrm{D}g(u)^*[\alpha]\rVert^2 = \bigl(\mathrm{D}g(u)^*\bigr)^{\dagger}[v],
\end{equation}
where $(\cdot)^{\dagger}$ denotes the Moore--Penrose pseudoinverse.
Substituting Eq.~(\ref{eq:alpha}) into Eq.~(\ref{eq:v_decompose}), we obtain a formula for the orthogonal projection from~$\mathcal{E}$ to~$\mathrm{T}_u \mathcal{M}$,
\begin{equation}
  \label{eq:proj}
  \mathrm{Proj}_u(v)  = v - \mathrm{D}g(u)^* \left[\bigl(\mathrm{D}g(u)^*\bigr)^{\dagger}[v]    \right].
\end{equation}

\hfill

\paragraph{Projected Neural Differential Equation.}
Projecting the NDE in Eq.~(\ref{eq:nde}) using the projection operator defined in Eq.~(\ref{eq:proj}) gives the following \emph{Projected Neural Differential Equation} (PNDE),
\begin{equation}
  \label{eq:pnde}
  \dot u = \mathrm{Proj}_u (\bar f_{\theta}(u))  \in \text{T}_u\mathcal{M},
\end{equation}
which is a differential equation on the constraint manifold~$\mathcal{M}$.
\begin{proposition}
  Solutions to the PNDE~(\ref{eq:pnde}) with~$u_0 \in \mathcal{M}$ satisfy~$g(u(t)) =0$ for all~$t \geq 0$.
  \begin{proof} We note that
    \begin{equation}
      \frac{dg(u(t))}{dt} = \mathrm{D}g(u) [\dot u ]= \mathrm{D}g(u)  \left[ \mathrm{Proj}_u (\bar f_{\theta}(u)) \right] = 0
    \end{equation}
    since $\mathrm{Proj}_u (\bar f_{\theta}(u))  \in \text{T}_u\mathcal{M} = \ker \mathrm{D}g(u)$. 
    As a result, $g(u(t))$ remains constant in time and satisfies $g(u(t)) = g(u(0)) = 0$ for all~$t \geq 0$  since $u_0 \in \mathcal{M}$.
  \end{proof}
\end{proposition}
Using matrix notation for the Euclidean space~$\mathcal{E} = \mathbb{R}^n$, let~$G(u) \in \mathbb{R}^{m \times n}$ denote the Jacobian matrix of~$g$ at~$u$, with entries~$G_{ij}(u) = \partial g_i / \partial u_j$.
Under this identification, the differential acts as~$\mathrm{D}g(u)[v] = G(u) v$ and its adjoint as~$\mathrm{D}g(u)^*[\alpha] = G(u)^\top \alpha$.
Since~$G(u)$ has full row rank by assumption, the pseudoinverse simplifies to~$(G^\top)^\dagger = (GG^\top)^{-1}G$, and the projection operator in Eq.~(\ref{eq:proj}) takes the matrix form
\begin{equation}
  \label{eq:proj_matrix}
  \mathrm{Proj}_u(v) = \left( I - G(u)^\top \bigl(G(u) G(u)^\top\bigr)^{-1} G(u) \right) v.
\end{equation}

\hfill

\paragraph{Generalization to Riemannian Metrics.}
While the standard Euclidean inner product is mathematically straightforward and computationally efficient, some physical systems may require non-Euclidean geometries.
For example, in systems where state variables possess disparate characteristic scales, a weighted inner product can provide physically consistent scaling, ensuring that each variable contributes proportionally to the projection.
We focus on the Euclidean case in this paper and provide a derivation of the projection operator for arbitrary Riemannian metrics in Appendix~\ref{sec:appendix_riemannian}.

\subsection{Practical Implementation}
\label{sec:implementation}

\paragraph{Forward Pass.}
To evaluate Eq.~\eqref{eq:proj_matrix}, we first solve the linear system
\begin{equation}
  \label{eq:linear_system}
  G G^\top \lambda = G v
\end{equation}
for~$\lambda \in \mathbb{R}^m$.
Then, we simply evaluate the projected vector as
\begin{equation}
  \label{eq:proj_operator}
  \mathrm{Proj}_u(v) = v - G^\top \lambda.
\end{equation}
This approach avoids explicitly forming the matrix inverse and permits the use of highly optimized batched solvers. %
In particular, since the Gram matrix $G G^\top$ is $m \times m$ symmetric positive definite, we can use the Cholesky factorization $G G^\top = L L^\top$ to solve Eq.~(\ref{eq:linear_system}) efficiently.

Appropriate batched and GPU-accelerated routines are readily available in the \texttt{cuSOLVER} library, part of the standard NVIDIA CUDA toolkit.
In practice, we find that the equivalent routines in the \texttt{MAGMA} library\cite{magma} are slightly faster and we use them throughout this paper.

\paragraph{Backward Pass.}
To train the neural network parameters via gradient descent, we require derivatives of the projection operator.
In particular, we need to propagate gradients of projection outputs backwards to gradients of projection inputs.
Consider a scalar loss function~$\mathcal{L}$ and let~$\bar p = \nabla_p \mathcal{L}$ denote the gradient with respect to the projected vector~$p = v - G^\top \lambda$, where~$\lambda$ solves~$G G^\top \lambda = G v$.
We need the gradients~$\bar v = \nabla_v \mathcal{L}$ and~$\bar G = \nabla_G \mathcal{L}$.
Because existing automatic differentiation frameworks (such as \texttt{Zygote.jl}, which we use here) cannot differentiate calls to external solver libraries (e.g., \texttt{cuSOLVER} or \texttt{MAGMA}), we instead derive analytic expressions for $\bar v$ and $\bar G$.
We provide a detailed derivation in Appendix \ref{appendix:projection_gradients} and outline the key results here.

Firstly, the gradient with respect to~$v$ is given by
\begin{equation}
  \label{eq:grad_v}
  \bar v = \bar p - G^\top \mu,
\end{equation}
where~$\mu \in \mathbb{R}^m$ solves the adjoint system
\begin{equation}
  \label{eq:adjoint_system}
  G G^\top \mu = G \bar p.
\end{equation}
Note that this has the same form as the forward pass in Eq.~(\ref{eq:proj_operator}), except the projection is applied to~$\bar p$ instead of~$v$.
Consequently, we can reuse the Cholesky factorization~$G G^\top = L L^\top$ from the forward pass to solve Eq.~(\ref{eq:adjoint_system}) efficiently, reducing the evaluation of Eq.~(\ref{eq:grad_v}) to a single $m \times m$ triangular solve followed by a matrix-vector multiplication.

Deriving the gradient with respect to~$G$ is more involved, as~$G$ appears both explicitly in~$p = v - G^\top \lambda$ and implicitly via $\lambda$ from the linear system~$G G^\top \lambda = G v$.
However, applying the chain rule to each contribution and simplifying yields the surprisingly compact form
\begin{equation}
  \label{eq:grad_G}
  \bar G = -\mu p^\top - \lambda \bar v^\top.
\end{equation}
Note that all quantities on the right-hand side of Eq.~(\ref{eq:grad_G}) have already been computed: $p$ and $\lambda$ from the forward pass, and $\bar v$ and $\mu$ from Eq.~(\ref{eq:grad_v}).
Thus, by caching and reusing these quantities, the entire backward pass reduces to matrix-vector multiplications and a single $m \times m$ triangular solve, thereby adding minimal overhead compared to the forward pass.
\paragraph{Alternative Approaches.}
The approach outlined so far is computationally efficient but may not be optimal when $G$ is ill-conditioned, since forming the Gram matrix squares the condition number, that is, $\kappa(GG^\top) = \kappa(G)^2$.
In this case, we can instead use QR decomposition to solve Eq.~(\ref{eq:linear_system}) without explicitly forming the Gram matrix.
See Appendix~\ref{appendix:qr_projection} for the forward and backward passes in this case.

\paragraph{Computational Complexity.}
Appendix~\ref{appendix:complexity} summarizes the complexity of each operation for both the Cholesky- and QR-based implementations outlined above.
In each case, the total complexity of the forward pass is $O(m^2 n)$, although the Cholesky-based approach has a smaller constant factor.
In both implementations, the backward pass complexity is reduced to $O(mn)$ by reusing the factorizations from the forward pass, allowing for highly efficient training.

Crucially, the linear scaling with system dimension~$n$ allows our approach to readily handle high-dimensional state spaces.
The primary computational bottleneck is instead determined by the constraint dimension~$m$, making the method most efficient for systems where constraints are relatively sparse compared to the state variables ($m \ll n$).
Nonetheless, the quadratic scaling with $m$ is mitigated by the extreme efficiency of dense linear algebra on modern accelerators, and we observe competitive wall-clock times even for large values of $m$ (Section~\ref{sec:computational_performance}).
\section{Related Work}
\label{sec:related_work}
\paragraph{Hard Constraints.}
A number of projection-based approaches have been proposed to constrain the outputs of deep learning models.
\citet{Harder2023} design constraint layers to ensure mass and energy conservation during downscaling (i.e., super-resolution) of climate model fields from low to high resolution.
\citet{Negiar2023} and \citet{chalapathi2024} add a differentiable implicit constraint layer for finding optimal linear combinations of learned basis functions that approximately satisfy differential constraints at collocation points.
\citet{jiang2020} and \citet{Duruisseaux2024Projection} use a spectral projection layer to enforce linear differential constraints efficiently in Fourier space.
Hard constraints have also been used to take advantage of structured and well-understood dynamical systems, for example, the divergence-free property of incompressible fluid flows\citep{Richter2022,Mohan2023,Xing2024} and the symplectic structure of Hamiltonian systems.\cite{Lutter2018,greydanus_hamiltonian_2019,Zhong2020,Jin2020,BurbyHenon,cranmer_lagrangian_2020,Saemundsson2020,Valperga2022,Duruisseaux2023NPMap,DuruisseauxLieFVINs}

The work of~\citet{finzi_simplifying_2020} is closely related to ours.
They use explicit algebraic constraints to enable the learning of Hamiltonian dynamics in Cartesian coordinates, where the Hamiltonian is ``simpler'' and therefore easier to learn than in the original, generalized coordinates.
They add constraints to the Hamiltonian via Lagrange multipliers, and use a variational approach to derive the corresponding constrained equations of motion.
The result is the standard Hamiltonian equations of motion, multiplied by a projection operator that enforces the constraints in a manner that is consistent with the underlying Hamiltonian structure.
While derived from an entirely different perspective, our approach can be understood as a generalization of their method beyond the idealized Hamiltonian setting to a more general class of NDEs.
In this paper, for example, we include tests on a damped pendulum system, which is not Hamiltonian.

\paragraph{Stabilized Neural Differential Equations.}
\citet{white_stabilized_2023} constrain NDE solutions using stabilized neural differential equations (SNDEs), given by
\begin{equation}
  \label{snde}
  \dot u = f_{\theta}(u) - \gamma F(u) g(u),
\end{equation}
where $\gamma \ge 0$ is a scalar, $F : \sR^n \to \sR^{n \times m}$ is a stabilization matrix, and $g: \sR^n \to \sR^m$ is the same constraint function we defined in Eq.~(\ref{manifold}).
Under mild conditions on $\gamma$ and $F$, the stabilized NDE~(\ref{snde}) admits all solutions of the original NDE (\ref{eq:nde}) while rendering the constraint manifold provably asymptotically stable.
In other words, the stabilized NDE (\ref{snde}) is able to learn the correct ground truth dynamics while guaranteeing that the constraints are satisfied.

To understand how SNDEs relate to our proposed method, we first provide the intuition behind Eq.~(\ref{snde}).
The stabilization term (i.e., the second term on the right-hand side) vanishes on the manifold, since $g(u) = 0$ there by definition.
Hence, the stabilization is only ``activated'' when a trajectory leaves the constraint manifold, that is, when the constraints have already been violated.
The effect of the stabilization is then to ``nudge'' such a trajectory ``back toward $\mathcal{M}$,'' i.e., back toward satisfying the constraints.
The strength of the nudge is determined by the stabilization parameter~$\gamma$; larger values of~$\gamma$ lead to a stronger nudge and stricter enforcement of the constraints.
However, excessively large values of~$\gamma$ may cause the system to become stiff, requiring the numerical solver to take smaller steps.

The key difference with our method is that SNDEs continuously correct violations of the constraints, while our method, by ensuring that velocities are always tangent to the constraint manifold, prevents such violations from occurring in the first place.
Accordingly, we can expect our method to yield a stricter enforcement of the constraints.
In Appendix~\ref{appendix:snde_limit}, we further show that the SNDE in Eq.~(\ref{snde}) reduces in the limit $\gamma \to \infty$ to the PNDE in Eq.~(\ref{eq:pnde}).
We will use SNDEs as a baseline for comparison throughout this paper.

\paragraph{Continuous Normalizing Flows on Riemannian Manifolds.}
A large body of related work has studied neural ODEs on Riemannian manifolds, primarily in the context of continuous normalizing flows on non-Euclidean geometries.
In contrast to this paper, they typically deal with prototypical Riemannian manifolds, such as spheres and tori, and are not easily adapted to more general manifolds.
For example, the approach of~\citet{lou_neural_2020} requires a local chart for the manifold, the derivation of which is likely to be analytically intractable for embedded submanifolds with arbitrary and nonlinear local defining functions.
Similarly, the approaches of~\citet{bose_latent_2020} and~\citet{rezende_normalizing_2020} require an expression for the exponential map, while the approach of~\citet{mathieu_riemannian_2020} uses a custom geodesic distance layer, none of which are easily applied in our setting.

\paragraph{Optimization and Numerical Integrators on Manifolds.}
The use of projections is widespread in optimization on manifolds,\cite{Absil2008,boumal_introduction_2024} where it is used to project iterates to the constraint manifold and\slash or gradients onto appropriate tangent spaces (as in our approach).
The idea of projecting gradients onto tangent spaces of the constraint set dates back several decades,\cite{Luenberger72,Luenberger73} and is still commonly used.
For instance, in Riemannian optimization, Euclidean gradients are typically replaced by Riemannian gradients, which are their orthogonal projections onto tangent spaces.
Explicit formulas and efficient projection algorithms have been derived and used for manifolds of greater practical interest.\cite{Edelman1998,Manton2002,Absil2008,Wen2010,Absil2012,Hauswirth2016,Sra2016,Oviedo2019,Oviedo2021}
Where these methods differ crucially from our approach is that they typically seek to constrain the weights or optimization variables themselves to a given matrix manifold, while our aim is to constrain trajectories of an NDE, the weights of which are not themselves directly constrained.
Regarding numerical integration on manifolds, the available literature is much less extensive,\cite{Hairer2011} and most approaches restrict themselves to Lie groups (very structured Riemannian manifolds) and Hamiltonian or mechanical systems, in order to obtain practical algorithms.\cite{Iserles2000,Christiansen2011,Celledoni2014,Celledoni2022}
Some of these approaches are based on projection, similarly to optimization on manifolds, while others use local parametrizations or Lagrange multipliers.
Riemannian optimization algorithms have also been obtained by simulating dynamical systems using numerical integrators constrained to the relevant Riemannian manifolds.\cite{alimisis2020,Tao2020,Lee2021,Franca2021,Duruisseaux2022Constrained,Duruisseaux2022Projection,Duruisseaux2022Riemannian,Duruisseaux2022Lagrangian}

\paragraph{Learned Constraints.}
Methods have been introduced that learn and enforce constraints from data.
The ``Constants Of Motion nETwork'' (COMET)\cite{kasim_constants_2022} imposes orthogonality between the learned vector field and discovered conserved quantities.
Similarly, \citet{yang_learning_2020} propose a neural projection operator to learn underlying physical constraints from position data, which are then enforced via an iterative correction scheme.
More recently, FINDE\cite{matsubara_finde_2023} utilizes projection methods to find and preserve unknown first integrals.
However, all of these methods differ fundamentally from ours in their treatment of constraints: while they aim to \emph{discover} hidden symmetries or constraints, our approach addresses the challenge of enforcing \emph{known} physical laws.
This distinction has significant implications for computational efficiency; because discovery-based approaches parameterize the constraints with a neural network, they incur substantial gradient overhead to calculate the Jacobian of the constraints network at every function evaluation.
In contrast, our method avoids this overhead by leveraging the analytic Jacobian of the known constraints, enabling a highly optimized and exact projection (Section~\ref{sec:implementation}).
In most practical applications, especially regarding physical systems, the constraints are a priori known and fixed, and the optimized routines presented in this paper are thus of significant practical utility.

\section{Experiments}
\label{sec:experiments}
We now apply and evaluate our proposed approach on several test systems, including prototypical dynamical systems and predator-prey models subject to conservation laws, strongly chaotic pendulums with kinematic constraints, and a complex power grid model.
We benchmark our approach against both unconstrained and constrained baselines, specifically vanilla neural differential equations and stabilized neural differential equations, respectively.
We evaluate each model's predicted trajectory $\hat u(t)$ versus the ground truth trajectory $u(t)$ using the relative error $\| u(t) - \hat u(t) \|_2 / \| u(t)\|_2$.
For the constraint error we report $\|g(\hat u(t))\|_2$, which is zero on the constraint manifold.
All test statistics in this section are averaged over 100 unseen initial conditions.
We provide full details of the data generation, model setup, and training procedure in Appendices~\ref{app:data} and \ref{app:training}, and include illustrations of individual test trajectories in Appendix~\ref{app:supplementary_figures}.
Analytic constraint Jacobians for all systems are given in Appendix~\ref{appendix:jacobians}.

\subsection{Conservation Laws}
\subsubsection{2-D Particle in a Quartic Potential}
We begin with a particle moving in two dimensions in a radially symmetric quartic potential given by
\begin{equation}
  \label{eq:quartic_potential}
  V(x,y) = \frac{1}{4}(x^2 + y^2)^2 = \frac{r^4}{4},
\end{equation}
where $r = \sqrt{x^2 + y^2}$.
For a particle with unit mass, the equations of motion are
\begin{equation}
  \label{eq:quartic}
  \ddot{x} = -x(x^2 + y^2), \qquad \ddot{y} = -y(x^2 + y^2).
\end{equation}
Eq.~(\ref{eq:quartic}) produces bounded, quasi-periodic orbits that precess and do not close.

Converting to first-order form with state $u = (x, y, \dot{x}, \dot{y}) \in \sR^4$, the system conserves both energy,
\begin{equation}
  \label{eq:quartic_energy}
  E(u) = \frac{1}{2}(\dot{x}^2 + \dot{y}^2) + \frac{1}{4}(x^2 + y^2)^2,
\end{equation}
and angular momentum,
\begin{equation}
  \label{eq:quartic_angular}
  L(u) = x \dot{y} - y \dot{x}.
\end{equation}
Knowledge of both conservation laws yields an integrable system.
Instead, as a proof of concept, we suppose incomplete knowledge of the system and enforce only conservation of energy via the constraint
\begin{equation}
  g(u) = E(u) - E(u_0) = 0,
\end{equation}
where $g:\sR^{4} \to \sR^{1}$ and $u_0$ is the initial condition.

Despite the relative simplicity of this system, its precessing orbits pose a challenge for neural network approximation~(Fig.~\ref{fig:combined_qp_lv}).
The unconstrained NDE exhibits rapidly growing errors, with some trajectories spiraling inward, while others quickly accumulate errors in the precession angle~(Fig.~\ref{fig:quartic_potential_phase}).
Of the constrained approaches, PNDEs predict the state most accurately while enforcing the constraints to near machine precision.
\subsubsection{Lotka--Volterra System}
We next consider the classic Lotka--Volterra predator-prey model, describing the interaction between a prey population $x$ and a predator population $y$.
The dynamics are governed by the coupled differential equations
\begin{equation}
  \label{eq:lotka_volterra}
  \dot{x} = \alpha x - \beta xy, \qquad \dot{y} = \delta xy - \gamma y,
\end{equation}
where $\alpha, \beta, \gamma, \delta > 0$ are interaction parameters.
For any positive initial condition, the populations oscillate periodically and obey a conservation law given by
\begin{equation}
  \label{eq:lv_conserved}
  V(x,y) = \delta x - \gamma \ln x + \beta y - \alpha \ln y.
\end{equation}
To evaluate our method, we enforce this conservation law as the constraint
\begin{equation}
  g(u) = V(u) - V(u_0) = 0,
\end{equation}
where $u=(x,y)$.
While this constraint does not correspond to a well-defined physical property of the system, such as conservation of energy, this system nonetheless presents a distinct challenge compared to the other systems in this paper: the domain is restricted to the positive quadrant ($x,y > 0$) and the constraint is non-polynomial (involving logarithms).
As a result, the Jacobian of the constraint, given by
\begin{equation}
  G(x,y) =
  \begin{pmatrix}
    \delta - \gamma/x & \beta - \alpha/y
  \end{pmatrix},
\end{equation}
diverges as either population approaches zero.
For SNDEs, this causes training to become unstable when the stabilization term is evaluated near the domain boundary, where it diverges, and we were unable to train SNDEs on this system.
In contrast, PNDEs avoid this issue entirely by constraining trajectories to the manifold exactly, meaning that the Jacobian is never evaluated near the domain boundary.

\begin{figure}
  \centering
  \includegraphics[width=\columnwidth]{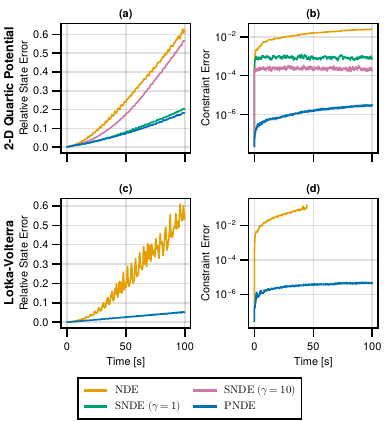}
  \caption{Comparison of NDE, SNDE, and PNDE on two benchmark systems.
    \textbf{Top row}: 2-D quartic potential showing (a) relative state error and (b) energy conservation error.
    \textbf{Bottom row}: Lotka--Volterra system showing (c) relative state error and (d) conserved quantity error.
    On both systems, PNDEs clearly outperform the baselines with respect to both constraint error and state error, achieving constraint errors near machine precision and better state accuracy than NDEs and SNDEs.
    SNDEs become unstable during training on the Lotka--Volterra system when the constraint Jacobian is evaluated near the domain boundary, where it diverges.
  }
  \label{fig:combined_qp_lv}
\end{figure}

As shown in Fig.~\ref{fig:combined_qp_lv}, the unconstrained NDE rapidly accumulates errors, leaving PNDE as the only method able to learn the dynamics in this setting.

\subsection{Simplifying Complex Learning Tasks Using Explicit Constraints}
\begin{figure}[t]
  \centering
\usetikzlibrary{calc,arrows.meta}

\begin{tikzpicture}[scale=2.2]

\draw [line width=2pt] (-0.5,0) -- (0.5,0);

\def\l{0.5} %
\def\thetaOne{40} %
\def\thetaTwo{-60} %
\def\thetaThree{35} %
\def\thetaFour{-45} %

\coordinate (O) at (0,0);

\def\phiOne{\thetaOne}
\def\phiTwo{\thetaTwo}
\def\phiThree{\thetaThree}
\def\phiFour{\thetaFour}

\coordinate (P1) at ($(O)+(270-\phiOne:\l)$);
\coordinate (P2) at ($(P1)+(270-\phiTwo:\l)$);
\coordinate (P3) at ($(P2)+(270-\phiThree:\l)$);
\coordinate (P4) at ($(P3)+(270-\phiFour:\l)$);

\draw [line width=1.5pt] (O) -- (P1);
\draw [line width=1.5pt] (P1) -- (P2);
\draw [line width=1.5pt] (P2) -- (P3);
\draw [line width=1.5pt] (P3) -- (P4);

\filldraw [black] (P1) circle (2.5pt);
\filldraw [black] (P2) circle (2.5pt);
\filldraw [black] (P3) circle (2.5pt);
\filldraw [black] (P4) circle (2.5pt);

\node [above] at (O) {$O$};

\draw [gray, dashed, thick] (O) -- ++(270:0.45);
\draw [gray, dashed, thick] (P1) -- ++(270:0.45);
\draw [gray, dashed, thick] (P2) -- ++(270:0.45);
\draw [gray, dashed, thick] (P3) -- ++(270:0.45);

\coordinate (T1) at ($(O)+(251:0.27)$);
\coordinate (T2) at ($(P1)+(302:0.23)$);
\coordinate (T3) at ($(P2)+(253:0.33)$);
\coordinate (T4) at ($(P3)+(294:0.27)$);

\node at (T1) {$\theta_1$};
\node at (T2) {$\theta_2$};
\node at (T3) {$\theta_3$};
\node at (T4) {$\theta_4$};

\end{tikzpicture}   \caption{Illustration of a $4$-pendulum.}
  \label{fig:pendulum}
\end{figure}
We now consider a damped $N$-pendulum system, consisting of $N$ simple pendulums connected end to end.
$N=1$ corresponds to the canonical simple pendulum, $N=2$ yields the well-known chaotic double pendulum, and higher values of $N$ yield systems with progressively more complex and chaotic dynamics (Fig.~\ref{fig:pendulum} shows $N=4$).
For realism, we include damping proportional to the relative angular velocity of each pendulum bob; this could represent, for example, friction in the hinges of the pendulum.

Assuming that each bob has a mass of \qty{1}{\kg} and each arm has a length of \qty{1}{\m}, the equations of motion of the $i$-th bob are given by
\begin{equation}
  \label{eq:n_pendulum}
  \begin{aligned}
    0  \ = \  &     \sum_{j=1}^{N}a(i,j)\bigl[ \ddot{\theta}_j \cos(\theta_i - \theta_j) + \dot \theta_j^2 \sin(\theta_i - \theta_j)\bigr] \\ & \quad +b(\dot\theta_i - \dot\theta_{i-1}) + (N-i+1) \mathfrak{g} \sin\theta_i,
  \end{aligned}
\end{equation}
where $\theta_i$ is the angle that the $i$-th arm makes with the vertical, $b = 0.1$ is the coefficient of friction, $a(i,j) = N - \max(i,j) + 1$, and $\mathfrak{g} = \qty{9.81}{m.s^{-2}}$ is acceleration due to gravity.

Given observations of this system, a reasonable approach would be to learn an NDE directly approximating Eq.~(\ref{eq:n_pendulum}), which is in generalized coordinates.
However, it is clear from Eq.~(\ref{eq:n_pendulum}) that the equations of motion will become increasingly complex as $N$ increases.
Alternatively, and analogously to the approach taken by previous work for Lagrangian and Hamiltonian systems,\cite{finzi_simplifying_2020} we can first convert to Cartesian coordinates, where the equations of motion may be easier to learn.
However, after converting to Cartesian coordinates, the length of each pendulum arm is no longer constant by construction, as is the case in generalized coordinates.
Hence, to enable learning in the ``easier'' Cartesian coordinates, we must \emph{explicitly} enforce that the length of each arm remains constant.

Given the position $q_i = (x_i, y_i)$ of the $i$-th bob in Cartesian coordinates, we can write down $N$ holonomic constraints on the system,
\begin{equation}
  \label{eq:holonomic}
  g^{(pos)}_i(q) =  \|q_i - q_{i-1}\|^2 -1 = 0,  \quad  i = 1, \ldots, N ,
\end{equation}
where $q = (q_1, \ldots, q_N)$ and $q_0 = (0,0)$ is the origin.
We differentiate these constraints to obtain $N$ additional, conjugate constraints on the velocity,
\begin{equation}
  \label{eq:non_holonomic}
  g^{(vel)}_i(q, \dot q) = (q_i - q_{i-1}) \cdot (\dot{q}_i - \dot{q}_{i-1}) = 0,  \quad  i = 1, \ldots, N.
\end{equation}
Collecting the position and velocity constraints into a vector, we obtain $2N$ constraints for the $N$-pendulum in Cartesian coordinates:
\begin{equation}
  g(u)  = \left[g^{(pos)}_1 , \ldots, g^{(pos)}_N, \ g^{(vel)}_1 , \ldots,  g^{(vel)}_N \right]  = 0,
\end{equation}
where $u = (q, \dot q)$ and $g:\sR^{4N} \to \sR^{2N}$.
We note again that the generalized coordinates of Eq.~(\ref{eq:n_pendulum}) satisfy these constraints automatically.

\begin{figure}[t]
  \centering
  \includegraphics[width=\columnwidth]{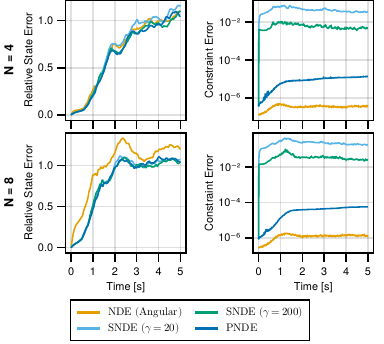}
  \caption{Comparison of NDE, SNDE, and PNDE on $N$-pendulum systems ($N = 4, 8$).
    \textbf{Left}: Relative state error over time.
    \textbf{Right}: Constraint violation $\|g(u)\|$ over time.
    NDE operates in constraint-satisfying generalized coordinates, while SNDE and PNDE operate in Cartesian coordinates with constraints explicitly enforced.
    The benefits of learning in Cartesian coordinates with constraints become pronounced as $N$ is increased.
    PNDEs enforce the constraints several orders of magnitude more accurately than SNDEs.
  }
  \label{fig:n_pendulum}
\end{figure}

Fig.~\ref{fig:n_pendulum} demonstrates that learning constrained dynamics in Cartesian coordinates produces better predictions than learning the same dynamics expressed in generalized coordinates.
The difference in performance becomes especially pronounced as $N$ is increased and the dynamics in generalized coordinates become more complex.
Among constrained models, PNDEs enforce the constraints several orders of magnitude more accurately than SNDEs and do not require tuning of the stabilization parameter~$\gamma$.
\subsection{Power Grids}
\begin{figure}[t]
  \centering
  \includegraphics[width=0.48\textwidth]{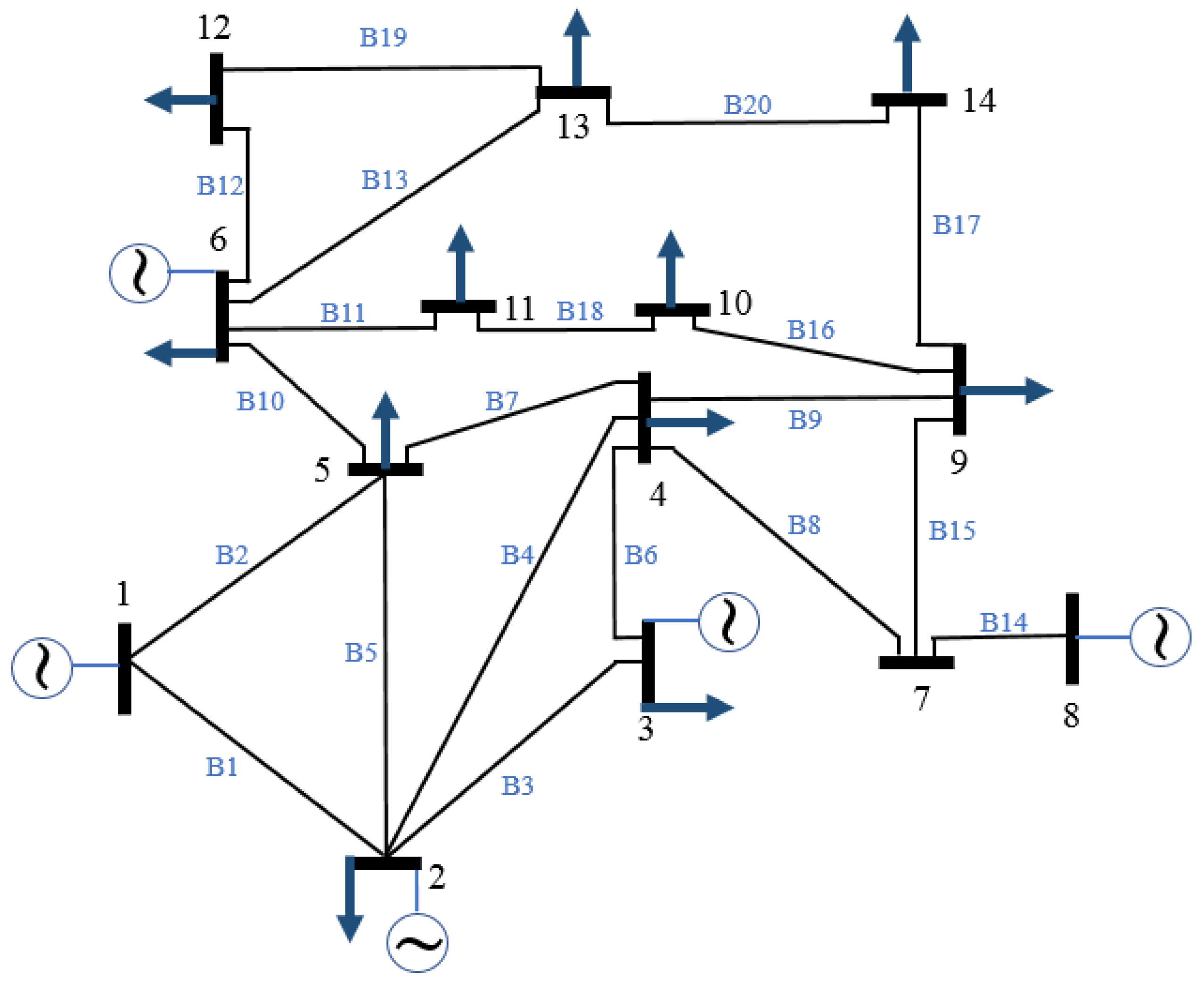}
  \caption{
    The IEEE 14-bus power grid system.
    Circles represent generators and arrows represent consumers.
    Figure reproduced with permission from~\citet{leon_quadratically_2020}.
  }
  \label{fig:ieee14bus}
\end{figure}

The IEEE test systems are standardized power system models widely used in electrical engineering research and education.\cite{christie_power_1999}
Representing key features of power grids such as topologies and operating conditions, these models enable researchers to benchmark algorithms and offer a common framework for analyzing power flow, stability, and reliability in electrical networks.
We study the dynamics of the IEEE 14-bus system (see Fig.~\ref{fig:ieee14bus}), adapted to represent future grid dynamics by replacing conventional generators with renewable energy sources.
Our renewables-based grids consist of three types of buses, or nodes: (1) renewable energy sources, (2) loads (that is, consumers), and (3) a slack bus (a constant-voltage bus representing, for example, a large power plant or a connection to another grid).

The state of each bus is determined by a complex voltage.
In this experiment, we treat the real and imaginary parts separately, so a grid with $N$ buses has $2N$ real-valued state variables.
Normally, a power grid operates at a steady-state equilibrium, called its \emph{operation point}.
We apply random perturbations (see Appendix~\ref{app:data}) and learn the dynamics from the resulting transients as the grid returns to its operation point.

The complex power at each PQ bus~$j$ is set to a fixed, constant value given by
\begin{equation}
  \label{eq:pq_bus}
  P_j +  \mathrm{i}Q_j = v_j \cdot i_j^*,
\end{equation}
where $P_j$ and $Q_j$ are the fixed active and reactive power, respectively, $v_j$ is the complex voltage, and $i_j$ is the complex current.\cite{machowski_power_2008}
These fixed power values will be the constraints in our experiments.
The complex currents in Eq.~(\ref{eq:pq_bus}) are determined by $i = \mathrm{LY} \cdot v$, where the admittance Laplacian $\mathrm{LY} \in \mathbb{C}^{N \times N}$ encodes the known line admittances.
Given $N$ total buses and $M$ PQ buses, the $2M$ constraints (one real and one imaginary for each PQ bus) are given by
\begin{equation}
  g(u) =
  \begin{bmatrix}
    P_1 - \Real{(v_1 \cdot i_1^*)} \\
    Q_1 - \Imag{(v_1 \cdot i_1^*)} \\
    \vdots \\
    P_M - \Real{(v_M \cdot i_M^*)} \\
    Q_M - \Imag{(v_M \cdot i_M^*)}
  \end{bmatrix}
  = 0,
\end{equation}
where $u = \left(\Real{(v_1)}, \Imag{(v_1)}, \ldots, \Real{(v_N)}, \Imag{(v_N)}\right) \in \sR^{2N}$ and $g: \sR^{2N} \to \sR^{2M}$.
For the IEEE 14-bus system, $N=14$ and $M=9$, so we learn the dynamics of 28 voltages subject to 18 constraints.

\begin{figure}[t]
  \centering
  \includegraphics[width=\columnwidth]{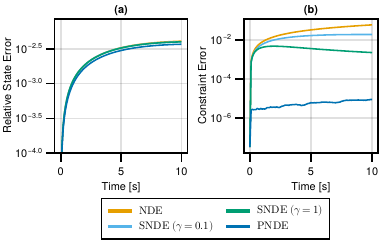}
  \caption{
    Comparison of NDE, SNDE, and PNDE for the IEEE 14-bus power grid system.
    (a)~Mean relative state error and (b)~mean constraint error over time on 100 test trajectories.
    All models produce similar state predictions, with SNDEs indistinguishable from the NDE.
    Stabilized NDEs (SNDE) with $\gamma \in \{0.1, 1\}$ reduce constraint drift compared to the unconstrained NDE but do not achieve exact constraint satisfaction.
    PNDEs enforce the constraints several orders of magnitude more accurately than SNDEs.
  }
  \label{fig:ieee14bus_comparison}
\end{figure}

As shown in Fig.~\ref{fig:ieee14bus_comparison}, all models produce similar state predictions on the test set, with SNDEs indistinguishable from the NDE.
However, unconstrained NDEs and SNDEs incur large errors in the PQ-bus power constraint.
For SNDEs, increasing the stabilization hyperparameter $\gamma$ enforces the constraints more accurately, but larger values lead to stiffness and, ultimately, prohibitively expensive simulations (see timings in Table~\ref{tab:times}).
Meanwhile, PNDEs enforce the constraints several orders of magnitude more accurately than the best SNDEs, without stiffness.

\begin{figure}[t]
  \centering
  \includegraphics[width=\columnwidth]{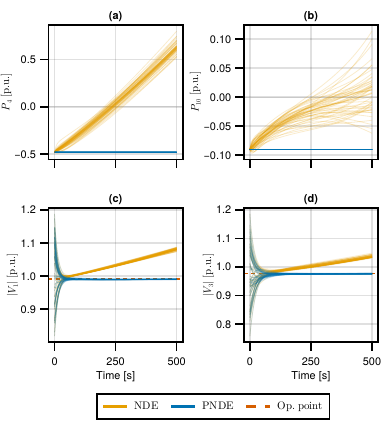}
  \caption{
    Long-horizon rollouts on the IEEE 14-bus system with 50 test initial conditions.
    \textbf{Top row}: Active power injection at PQ buses~4 and~10: the NDE trajectories drift upward and eventually turn positive (unphysical for load buses, implying that they spontaneously generate power), while PNDE trajectories remain at their initial values.
    \textbf{Bottom row}: Voltage magnitude at PV buses~1 and~3: PNDE trajectories converge to the operation point, whereas NDE trajectories do not converge.
    We plot the remaining buses in \cref{fig:ieee14bus-long-rollout-supp-pq,fig:ieee14bus-long-rollout-supp-pv} in Appendix~\ref{app:supplementary_figures}.
  }
  \label{fig:ieee14bus-long-rollout}
\end{figure}

\Cref{fig:ieee14bus-long-rollout} compares PNDEs and NDEs on significantly longer rollouts of duration $500\,\mathrm{s}$.
For unconstrained NDEs, the active power injected at the PQ (load) buses drifts and even turns positive, implying that they spontaneously generate power---this behavior is clearly unphysical, emphasizing the need for constraints.
By contrast, for PNDEs, the active power injected at the PQ buses remains constant due to the constraints.
For the PV buses, we compare the predicted voltage magnitude to the system's operation point, to which it should converge.
We find that the PNDE trajectories converge almost exactly to the correct operation point, while NDE trajectories drift and do not converge at all.
It is remarkable that the PNDE trajectories predict the operation point so accurately, given the highly limited training data---just 100 trajectories of duration $10\,\mathrm{s}$ each.
For completeness, we plot the remaining buses in \cref{fig:ieee14bus-long-rollout-supp-pq,fig:ieee14bus-long-rollout-supp-pv} in Appendix~\ref{app:supplementary_figures}.

\subsection{Computational Performance}
\label{sec:computational_performance}
Table~\ref{tab:times} shows the training and inference times for each system and model.
Compared to unconstrained NDEs, PNDEs inevitably incur additional computation in order to enforce the constraints.
The time required to train PNDEs ranges from $1.15$--$2.58\times$ the time required for the unconstrained NDE.
During inference, the time required for a single forward pass of the PNDE, that is, a single evaluation of the MLP and the constraints, is $1.23$--$2.81\times$ that of the unconstrained NDE.

We also report the time required to solve each model for an actual trajectory.
Compared to benchmarking a single forward pass of the model, this approach also reflects differences in stiffness between models that may require the adaptive solver to take more steps.
The time required for the PNDEs ranges from $1.00$--$2.13\times$ the time required for the unconstrained NDE.

Compared to SNDEs, the PNDE is always at least as fast to train and faster for a single forward pass, even for low values of the stabilization parameter $\gamma$.
This is because the SNDE requires evaluating both the constraint function $g$ and its Jacobian $G$ at each step, whereas the PNDE only requires $G$.
The PNDE is likewise faster than the SNDE for full trajectory solves in all cases except the $8$-pendulum, where the adaptive solver requires more steps for the PNDE than for the SNDE at $\gamma = 20$.

The different values of $\gamma$ reflect the trade-off between constraint accuracy and computational cost that exists for SNDEs.
Large values of $\gamma$ enforce the constraints more accurately, but eventually lead to stiffness and significantly more expensive solves.
PNDEs, in contrast, enforce the constraints close to machine precision without stiffness.
\begin{table}[t]
  \caption{
    Training and inference times for each system and model, using a batch size of 1024 on a single NVIDIA H100 GPU.
    The forward pass time is the duration of a single forward pass of the model, that is, a single evaluation of the MLP and, where applicable, the constraints.
    The solve time is the duration to obtain a single solution of the model for an actual trajectory, where the duration of the trajectory for a given system is the same as the test trajectories used in the previous sections.
    Forward-pass and solve times report the minimum over repeated runs.
  }

  \label{tab:times}
  \begin{ruledtabular}
    \begin{tabular}{llrrr}
      System                                 & Model                 & Train & Fwd. Pass & Solve \\
      &                       & (min) & (ms)      & (ms)  \\[0.5ex]
      \hline\noalign{\vspace{0.5ex}}
      \multirow{4}{*}{\shortstack[l]{2-D Quartic\\Potential}} & NDE                   & 3.9   & 0.17      & 2163  \\
      & SNDE ($\gamma{=}1$)   & 4.5   & 0.50      & 2743  \\
      & SNDE ($\gamma{=}10$)  & 4.5   & 0.50      & 3908  \\
      & PNDE                  & 4.5   & 0.42      & 2531  \\[0.5ex]
      \hline\noalign{\vspace{0.5ex}}
      \multirow{3}{*}{Lotka--Volterra}       & NDE                   & 5.5   & 0.20      & 3201  \\
      & SNDE                  & N/A   & N/A       & N/A   \\
      & PNDE                  & 6.6   & 0.38      & 3199  \\[0.5ex]
      \hline\noalign{\vspace{0.5ex}}
      \multirow{4}{*}{$4$-Pendulum}          & NDE (Angular)         & 238   & 1.22      & 2235  \\
      & SNDE ($\gamma{=}20$)  & 703   & 1.68      & 3696  \\
      & SNDE ($\gamma{=}200$) & 755   & 1.70      & 6089  \\
      & PNDE                  & 607   & 1.50      & 3506  \\[0.5ex]
      \hline\noalign{\vspace{0.5ex}}
      \multirow{4}{*}{$8$-Pendulum}          & NDE (Angular)         & 320   & 1.23      & 3629  \\
      & SNDE ($\gamma{=}20$)  & 1005  & 1.95      & 5980  \\
      & SNDE ($\gamma{=}200$) & 890   & 1.90      & 8778  \\
      & PNDE                  & 825   & 1.75      & 7725  \\[0.5ex]
      \hline\noalign{\vspace{0.5ex}}
      \multirow{5}{*}{IEEE 14-Bus}           & NDE                   & 9.8   & 0.36      & 36    \\
      & SNDE ($\gamma{=}0.1$) & 25.5  & 1.22      & 75    \\
      & SNDE ($\gamma{=}1$)   & 23.3  & 1.22      & 83    \\
      & SNDE ($\gamma{=}10$)  & 22.8  & 1.23      & 245   \\
      & PNDE                  & 21.3  & 1.00      & 67    \\
    \end{tabular}
  \end{ruledtabular}
\end{table}
\section{Conclusion}
We have introduced projected neural differential equations (PNDEs), a method for enforcing hard algebraic constraints in neural differential equations by projecting the learned vector field onto the tangent space of the constraint manifold.
The approach provides provable constraint satisfaction to machine precision, requires no constraint-specific hyperparameters, and avoids the numerical stiffness that can affect stabilization-based alternatives.
Our projection method scales well with increasing system dimension and is applicable to any dynamical system subject to physical or kinematic constraints, thus covering a wide range of applications in science and engineering.
We derived analytic gradients of the projection operator, enabling efficient training with moderate overhead compared to unconstrained models.
Across experiments on chaotic dynamical systems, a predator-prey model, and a realistic power grid model, PNDEs consistently outperformed existing methods in constraint accuracy---by several orders of magnitude---while maintaining at least competitive and typically improved state prediction and computational efficiency.
We further showed that explicit constraints can enable effective learning in simpler coordinate systems, substantially improving generalization for high-dimensional chaotic systems.
These results suggest that PNDEs offer a principled and practical framework for incorporating known physical laws in neural differential equation models.

\hfill

\paragraph{Limitations.}
The computational overhead of projection---roughly $1.0$--$2.8\times$ for training and inference relative to an unconstrained NDE (Table~\ref{tab:times})---scales quadratically with the number of constraints~$m$.
For systems with many constraints, this may become a bottleneck, though in practice we find that the overhead remains moderate even for the largest constraint dimensions considered here.
The projection also assumes that the constraint Jacobian has full row rank everywhere on the constraint manifold, but this can fail at singular configurations where constraints become linearly dependent; the method does not currently handle such cases.
While PNDEs enforce constraints to machine precision, round-off errors may nonetheless lead to drift during long simulations beyond those studied here; in such cases, a periodic coordinate projection (see, e.g., \citet{Hairer2011}) may be applied to return the trajectory to the constraint manifold.

\hfill

\paragraph{Future Work.}
Although we focus on the Euclidean projection in this paper, the generalized projection operator derived in Appendix~\ref{sec:appendix_riemannian} suggests several practical directions.
While a projection based on the most general, state-dependent metric may be prohibitively expensive to evaluate, more physically motivated metrics---such as a constant diagonal weighting matrix for state variables with disparate scales or units, or the mass matrix of a mechanical system---may yield physically consistent projections at negligible additional cost.
Evaluating whether such metrics improve learning or long-horizon trajectory quality is a natural next step.
Finally, generalizing the projection approach to inequality constraints would broaden applicability to settings such as voltage bounds in power grid models, where one-sided constraints arise naturally.
\begin{acknowledgments}
  This work was funded by the Volkswagen Foundation.
  N.B. and M.G. acknowledge the ClimTip project that has received funding from the European Union’s Horizon Europe research and innovation programme under grant agreement No.~101137601; this is ClimTip contribution \#.
  F.H. acknowledges the eKI4Ds project, funded by the German Federal Ministry for Economy and Energy (03EI1092A).
  N.K. has received funding from the European Research Council (ERC) under the European Union’s Horizon Europe research and innovation programme (grant agreement No.~101221985).
  N.K. has been supported by the German Federal Ministry of Education and Research (Grant: 01IS24082).
  The authors gratefully acknowledge the European Regional Development Fund (ERDF), the German Federal Ministry of Education and Research, and the Land Brandenburg for supporting this project by providing resources on the high performance computer system at the Potsdam Institute for Climate Impact Research.
\end{acknowledgments}
\section*{Author Declarations}
\subsection*{Conflict of Interest}
The authors have no conflicts to disclose.
\section*{Data Availability}
The code required for generating the data and running the experiments in this study is available on GitHub at \url{https://github.com/white-alistair/Projected-Neural-Differential-Equations}.
\appendix

\section{Generalization to Riemannian Metrics}
\label{sec:appendix_riemannian}

In Section~\ref{sec:pnde}, we specialized the projection operator to the standard Euclidean inner product.
Here, we generalize the derivation to an ambient space~$\mathcal{E}$ equipped with a state-dependent Riemannian metric.
Let the inner product at~$u \in \mathcal{E}$ be defined by a smoothly varying symmetric positive-definite matrix~$M(u) \in \mathbb{R}^{n \times n}$, such that
\begin{equation}
  \langle v, w \rangle_{M(u)} = v^\top M(u) w
\end{equation}
for any tangent vectors~$v, w \in \text{T}_u \mathcal{E}$.

Let~$G(u) \in \mathbb{R}^{m \times n}$ denote the Jacobian matrix of the constraints~$g$ at~$u$.
The differential is given by~$\mathrm{D}g(u)[v] = G(u)v$. To find the orthogonal projection onto the tangent space~$\text{T}_u \mathcal{M} = \ker G(u)$ with respect to this metric, we must first determine the adjoint operator~$\mathrm{D}g(u)^*$ such that
\begin{equation}
  \langle G(u)v, \alpha \rangle_{\mathbb{R}^m} = \langle v, \mathrm{D}g(u)^*[\alpha] \rangle_{M(u)},
\end{equation}
where we assume the standard Euclidean inner product on the constraint space~$\mathbb{R}^m$.
Expanding both sides yields
\begin{equation}
  (G(u)v)^\top \alpha = v^\top M(u) \mathrm{D}g(u)^*[\alpha].
\end{equation}
Rearranging the left side as~$v^\top G(u)^\top \alpha$, we identify the adjoint under the Riemannian metric as
\begin{equation}
  \mathrm{D}g(u)^*[\alpha] = M(u)^{-1} G(u)^\top \alpha.
\end{equation}

Following the orthogonal decomposition in Eq.~(\ref{eq:v_decompose}), any vector~$v \in \mathcal{E}$ can be uniquely decomposed into a component parallel to~$\text{T}_u \mathcal{M}$ and a component orthogonal to it (with respect to~$M(u)$),
\begin{equation}
  v = \mathrm{Proj}_u^M(v) + M(u)^{-1} G(u)^\top \alpha.
\end{equation}
Because the projected vector must lie in the tangent space, it satisfies~$G(u) \mathrm{Proj}_u^M(v) = 0$.
Applying~$G(u)$ to both sides of the decomposition gives
\begin{equation}
  G(u) v = 0 + G(u) M(u)^{-1} G(u)^\top \alpha.
\end{equation}
Assuming~$G(u)$ has full row rank, the matrix~$G(u) M(u)^{-1} G(u)^\top$ is invertible.
Solving for~$\alpha$ yields
\begin{equation}
  \alpha = \bigl(G(u) M(u)^{-1} G(u)^\top\bigr)^{-1} G(u) v.
\end{equation}
Substituting~$\alpha$ back into the decomposition, we obtain the generalized projection operator,
\begin{equation}
  \mathrm{Proj}_u^M(v) = \left( I - M^{-1} G^\top \bigl(G M^{-1} G^\top\bigr)^{-1} G \right) v.
\end{equation}
Replacing the Euclidean projection in Eq.~(\ref{eq:pnde}) with~$\mathrm{Proj}_u^M$ yields a projected neural differential equation that is dynamically consistent with the chosen Riemannian geometry.

In practice, the choice of metric depends on the physical properties of the system being studied.
For state variables with disparate scales or units, a constant diagonal weighting matrix~$W$ is sufficient; since~$W^{-1}$ is trivially evaluated, this adds negligible cost compared to the Euclidean case and preserves the~$O(m^2 n)$ forward-pass complexity.
More general choices of $M$ require solving an~$n \times n$ system, increasing the forward-pass cost to~$O(n^3)$ and limiting their practical utility.

\section{Derivation of Projection Operator Gradients}
\label{appendix:projection_gradients}
We derive the gradients of the projection operator with respect to its inputs, enabling backpropagation through the projection layer.
\subsection{Setup}
The forward pass computes the projection
\begin{equation}
  \label{eq:forward_pass}
  p = v - G^\top \lambda, \quad \text{where } \lambda \text{ solves } G G^\top \lambda = G v.
\end{equation}
In matrix form, this is equivalent to $p = Pv$, where $P = I - G^\top (GG^\top)^{-1} G$ is the projection matrix.
Given a scalar loss $\mathcal{L}$ and its gradient $\bar p = \nabla_p \mathcal{L}$ with respect to the projection output $p$, we seek the gradients $\bar v = \nabla_v \mathcal{L}$ and $\bar G = \nabla_G \mathcal{L}$ with respect to the projection inputs $v$ and $G$.

We first state the necessary adjoint results from \citet{giles_collected_2008}:

\hfill

\paragraph{Matrix Multiplication (\citet[\S2.2]{giles_collected_2008}):}
If $C = AB$, the reverse mode sensitivities are:
\begin{equation}
  \label{eq:rule_mult}
  \bar A = \bar C B^\top, \quad \bar B = A^\top \bar C.
\end{equation}

\hfill

\paragraph{Matrix Inverse Product (\citet[\S2.3]{giles_collected_2008}):}
If $C = A^{-1}B$, the reverse mode sensitivities are:
\begin{equation}
  \label{eq:rule_inv_prod}
  \bar B = A^{-\top} \bar C, \quad \bar A = -\bar B C^\top.
\end{equation}
\subsection{Gradient with respect to~\texorpdfstring{$v$}{v}}
By the multiplication rule \eqref{eq:rule_mult}, if $p = Pv$ then $\bar v = P^\top \bar p$.
Since $P$ is symmetric, $\bar v = P \bar p$, which means we simply apply the projection to $\bar p$:
\begin{equation}
  \label{eq:grad_v_app}
  \bar v = \bar p - G^\top \mu,
\end{equation}
where $\mu$ solves the adjoint system $G G^\top \mu = G \bar p$.
Note that this has the same structure as the forward pass.

\subsection{Gradient with respect to~\texorpdfstring{$G$}{G}}
The matrix $G$ enters the projection \eqref{eq:forward_pass} in three places: (i) explicitly in $G^\top \lambda$, (ii) via $Gv$ in the right-hand side of $G G^\top \lambda = G v$, and (iii) via the Gram matrix $GG^\top$ in the left-hand side of $G G^\top \lambda = G v$.
We apply the chain rule to each contribution separately.

\paragraph{Contribution from $G^\top \lambda$.}
Let  $q = G^\top \lambda$.
By the multiplication rule \eqref{eq:rule_mult}, $\bar G^\top = \bar q \lambda^\top$, and hence
\begin{equation}
  \label{eq:g_lambda_q}
  \bar G = \lambda \bar q^\top.
\end{equation}
Substituting $q$ into \eqref{eq:forward_pass}, we have $p = v - q$ and, by the chain rule, $\bar q = -\bar p$.
Substituting $\bar q$ into \eqref{eq:g_lambda_q}, we get
\begin{equation}
  \bar G^{(1)} = -\lambda \bar p^\top.
\end{equation}

\hfill

\paragraph{Contribution from $Gv$.}
Let $A = G G^\top$ and $b = Gv$ so that $A \lambda = b$.
By the multiplication rule \eqref{eq:rule_mult},
\begin{equation}
  \label{eq:rule_mult_b}
  \bar G = \bar b v^\top.
\end{equation}
By the inverse product rule \eqref{eq:rule_inv_prod}, if $\lambda = A^{-1}b$, then
\begin{equation}
  \label{eq:rule_inv_prod_b}
  \bar b = A^{-\top} \bar \lambda.
\end{equation}
To find $\bar \lambda$, note that $\lambda$ affects $p$ through $G^\top \lambda$; we again apply the multiplication rule \eqref{eq:rule_mult} to \eqref{eq:forward_pass} to get
\begin{equation}
  \label{eq:rule_mult_lambda}
  \bar \lambda = G(-\bar p) = -G\bar p.
\end{equation}
Substituting \eqref{eq:rule_mult_lambda} into \eqref{eq:rule_inv_prod_b}, we get
\begin{equation}
  \bar b = A^{-\top} (-G\bar p) = -A^{-1} G\bar p,
\end{equation}
where we have also used the fact that $A$ is symmetric.
Observe that this is exactly $-\mu$, where $\mu$ solves the adjoint system $GG^\top \mu = G\bar p$ introduced in the previous section.
Hence,
\begin{equation}
  \bar G^{(2)} = \bar b v^\top = -\mu v^\top.
\end{equation}

\hfill

\paragraph{Contribution from $GG^\top$.}
The Gram matrix $A = GG^\top$ also affects $\lambda$ via the linear system $\lambda = A^{-1}b$.
By the inverse product rule \eqref{eq:rule_inv_prod},
\begin{equation}
  \label{eq:rule_inv_prod_A}
  \bar A = -\bar b \lambda^\top.
\end{equation}
Substituting $\bar b = -\mu$ (established above) into \eqref{eq:rule_inv_prod_A}, we get
\begin{equation}
  \label{eq:rule_inv_prod_A_b}
  \bar A = \mu \lambda^\top.
\end{equation}
Since $A = GG^\top$, we can apply the matrix product rule \eqref{eq:rule_mult} twice to get
\begin{equation}
  \label{eq:G_3_matrix_product}
  \bar G^{(3)} = \bar A G + \bar A^\top G.
\end{equation}
Substituting \eqref{eq:rule_inv_prod_A_b} into \eqref{eq:G_3_matrix_product},
\begin{equation}
  \label{eq:rule_mult_A}
  \bar G^{(3)} = (\mu \lambda^\top + \lambda \mu^\top) G.
\end{equation}

\hfill

\paragraph{Combined gradient.}
Summing all three contributions:
\begin{equation}
  \label{eq:combined_grad_G}
  \bar G = -\lambda \bar p^\top - \mu v^\top + (\mu \lambda^\top + \lambda \mu^\top) G.
\end{equation}
This expression can be simplified using two relations obtained by rearranging earlier equations:
\begin{itemize}
  \item From the forward pass \eqref{eq:forward_pass}: $v = p + G^\top \lambda$.
  \item From the gradient w.r.t.\ $v$ \eqref{eq:grad_v_app}: $\bar p = \bar v + G^\top \mu$.
\end{itemize}
Substituting these into \eqref{eq:combined_grad_G} and expanding:
\begin{align}
  \bar G & = -\lambda \bar p^\top - \mu v^\top + \mu \lambda^\top G + \lambda \mu^\top G \nonumber                                           \\
  & = -\lambda (\bar v + G^\top \mu)^\top - \mu (p + G^\top \lambda)^\top + \mu \lambda^\top G + \lambda \mu^\top G \nonumber         \\
  & = -\lambda \bar v^\top - \lambda \mu^\top G - \mu p^\top - \mu \lambda^\top G + \mu \lambda^\top G + \lambda \mu^\top G \nonumber \\
  & = -\lambda \bar v^\top - \mu p^\top.
\end{align}
The cross terms $\lambda \mu^\top G$ and $\mu \lambda^\top G$ cancel, yielding the compact form
\begin{equation}
  \bar G = -\mu p^\top - \lambda \bar v^\top.
\end{equation}
This final expression requires only quantities already computed: $p$ and $\lambda$ from the forward pass, and $\bar v$ and $\mu$ from the gradient with respect to $v$.
\section{QR-Based Projection Implementation}
\label{appendix:qr_projection}
The default implementation of the projection operator, described in Section~\ref{sec:implementation}, forms the Gram matrix $GG^\top$ and solves the resulting symmetric positive-definite system via Cholesky factorization.
While computationally efficient, this approach squares the condition number: $\kappa(GG^\top) = \kappa(G)^2$.
For ill-conditioned constraint Jacobians, this can lead to numerical difficulties.
Here we describe an alternative implementation, based on QR decomposition, that avoids forming the Gram matrix entirely.
\subsection{Forward Pass}
We want to solve the linear system $GG^\top \lambda = Gv$ for $\lambda$ without explicitly forming $GG^\top$.
Consider the thin QR factorization of the transpose of the constraint Jacobian,
\begin{equation}
  \label{eq:qr_factorization}
  G^\top = QR,
\end{equation}
where $Q \in \sR^{n \times m}$ has orthonormal columns and $R \in \sR^{m \times m}$ is upper triangular.
Since $G$ has full row rank by assumption, $R$ is nonsingular.
Substituting Eq.~\eqref{eq:qr_factorization} into the Gram matrix yields
\begin{equation}
  GG^\top = (QR)^\top (QR) = R^\top Q^\top Q R = R^\top R,
\end{equation}
where we have used the orthonormality of $Q$.
The linear system $GG^\top \lambda = Gv$ therefore becomes
\begin{equation}
  \label{eq:qr_system}
  R^\top R \, \lambda = Gv,
\end{equation}
which can be solved by two successive triangular solves:
\begin{enumerate}
  \item Solve $R^\top z = Gv$ for $z$ (forward substitution),
  \item Solve $R\lambda = z$ for $\lambda$ (back substitution).
\end{enumerate}
Having obtained $\lambda$, the projected vector is computed as before via $p = v - G^\top \lambda$.

The key advantage is that $R$ inherits the condition number of $G$ rather than its square: $\kappa(R) = \kappa(G)$.
This follows from the fact that $R$ and $G$ have the same singular values, since $Q$ is orthogonal and therefore norm-preserving.
\subsection{Backward Pass}
The backward pass proceeds analogously to the Cholesky-based derivation in Appendix~\ref{appendix:projection_gradients}.
The gradient with respect to $v$ is
\begin{equation}
  \bar v = \bar p - G^\top \mu,
\end{equation}
where $\mu$ solves $R^\top R \, \mu = G \bar p$ via the same two-step triangular solve as in the forward pass.
The gradient with respect to $G$ retains the same form,
\begin{equation}
  \bar G = -\mu \, p^\top - \lambda \, \bar v^\top,
\end{equation}
since the derivation in Appendix~\ref{appendix:projection_gradients} depends only on the structure of the linear system $GG^\top \lambda = Gv$ and not on the specific factorization used to solve it.
In particular, by caching $R$ from the forward pass, the backward pass requires only two additional triangular solves and two outer products.
\section{Computational Complexity}
\label{appendix:complexity}
Table~\ref{tab:complexity} compares the complexity of the Cholesky- and QR-based implementations.
In both cases, the forward pass has complexity $O(m^2 n)$, dominated by formation of the Gram matrix $GG^\top$ and the QR factorization, respectively.
However, the QR-based approach avoids one batched matrix-matrix multiplication (the formation of $GG^\top$), at the expense of an additional triangular solve and a larger constant factor in the QR factorization itself.
The backward pass complexity is identical for both approaches, since both reuse the factorization from the forward pass.
\begin{table}[h]
  \caption{Comparison of Cholesky and QR implementations, where $n$ is the state dimension and $m < n$ is the number of constraints.}
  \label{tab:complexity}
  \begin{ruledtabular}
    \begin{tabular}{lcc}
      Operation                    & Cholesky   & QR         \\[0.5ex]
      \hline\noalign{\vspace{0.5ex}}
      Gram matrix $GG^\top$        & $O(m^2 n)$ & ---        \\
      Factorization                & $O(m^3)$   & $O(m^2 n)$ \\
      Triangular solve(s)          & $O(m^2)$   & $O(m^2)$   \\
      Matrix-vector multiplication & $O(mn)$    & $O(mn)$    \\[0.5ex]
      \hline\noalign{\vspace{0.5ex}}
      Forward pass total           & $O(m^2 n)$ & $O(m^2 n)$ \\
      Backward pass total          & $O(mn)$    & $O(mn)$    \\
    \end{tabular}
  \end{ruledtabular}
\end{table}

\section{PNDEs as the Infinite-\texorpdfstring{$\gamma$}{gamma} Limit of SNDEs}
\label{appendix:snde_limit}

In this section, we demonstrate that the projected neural differential equation (PNDE) naturally emerges as the infinite-penalty limit of the stabilized neural differential equation (SNDE).
Consider the SNDE,
\begin{equation}
  \label{eq:generalized_snde}
  \dot u = f_\theta(u) - \gamma F(u) g(u),
\end{equation}
where $\gamma > 0$ is the penalty parameter and $F : \sR^n \to \sR^{n \times m}$ is the stabilization matrix.
We assume the matrix $G(u)F(u) \in \sR^{m \times m}$ is invertible for all $u \in \mathcal{M}$, where $G(u)$ is the constraint Jacobian.

Consider the constraint violation $v(t) = g(u(t)) \in \sR^m$.
Differentiating $v$ along solutions of Eq.~(\ref{eq:generalized_snde}), we get
\begin{equation}
  \label{eq:snde_violation_generalized}
  \dot v = G(u) \dot u = G(u) f_\theta(u) - \gamma G(u) F(u) v.
\end{equation}
To take the limit, we introduce the rescaled multiplier $\lambda := \gamma v$ and the small parameter $\epsilon := 1/\gamma$.
Eq.~(\ref{eq:snde_violation_generalized}) then takes the standard singularly perturbed form:
\begin{equation}
  \label{eq:snde_singular_perturbation}
  \epsilon \dot \lambda = G(u) f_\theta(u) - G(u) F(u) \lambda.
\end{equation}
For each frozen $u$, the fast subsystem in Eq.~(\ref{eq:snde_singular_perturbation}) is a linear ODE in $\lambda$.
Because $G(u)F(u)$ is invertible, $\lambda$ relaxes on the fast timescale $\epsilon$ towards its quasi-steady state:
\begin{equation}
  \label{eq:snde_qss_generalized}
  \lambda^\star(u) = \bigl(G(u)F(u)\bigr)^{-1} G(u) f_\theta(u).
\end{equation}
As $\epsilon \to 0$ ($\gamma \to \infty$), this relaxation becomes instantaneous relative to the slow dynamics of $u$.
For initial conditions $u_0 \in \mathcal{M}$, the state $u$ starts exactly on the constraint manifold, though $\lambda$ undergoes a brief boundary layer of duration $O(\epsilon)$ as it transitions from $\lambda(0) = \gamma g(u_0) = 0$ to its quasi-steady state $\lambda^\star(u_0)$.
By Tikhonov's theorem, this transient vanishes in the limit $\epsilon \to 0$, ensuring $\lambda \to \lambda^\star(u)$ pointwise for $t > 0$.

Substituting $\lambda^\star(u)$ back into the generalized SNDE in place of $\gamma g(u)$ gives the limiting equation of motion on the manifold:
\begin{equation}
  \label{eq:snde_limit_oblique}
  \dot u = f_\theta(u) - F(u) \bigl(G(u)F(u)\bigr)^{-1} G(u) f_\theta(u) = P_F(u) f_\theta(u),
\end{equation}
where $P_F(u) := I - F(u) \bigl(G(u)F(u)\bigr)^{-1} G(u)$. We note that $P_F(u)$ is idempotent ($P_F(u)^2 = P_F(u)$) and $G(u)P_F(u) = 0$, meaning it acts as a projection operator onto the tangent space $\mathrm{T}_u \mathcal{M} = \ker G(u)$.

The derivation above reveals that in the infinite-$\gamma$ limit, \emph{every} valid SNDE acts as an oblique projector, pushing the vector field onto the tangent space along the column space of $F(u)$.
To obtain an \emph{orthogonal} projection, the image of $F(u)$ must be the orthogonal complement to the tangent space.
This immediately implies that the columns of $F(u)$ must span the same subspace as the constraint gradients; canonical choices are $F(u) = G(u)^\top$ and $F(u) = G(u)^\dagger$.
Substituting either choice into Eq.~(\ref{eq:snde_limit_oblique}) yields:
\begin{equation}
  \dot u = \left( I - G(u)^\top \bigl( G(u) G(u)^\top \bigr)^{-1} G(u) \right) f_\theta(u),
\end{equation}
which recovers the PNDE of Eq.~(\ref{eq:pnde}) with the orthogonal projection operator of Eq.~(\ref{eq:proj_matrix}).
\section{Datasets}
\label{app:data}

For each system, we generate ground truth trajectories from random initial conditions.
The exact number of initial conditions used for training, validation, and testing, as well as the duration of each trajectory and the timestep used, is determined based on the individual characteristics of each system, and summarized in Table~\ref{tab:data-hyperpar}.

For training and validation, each trajectory is split into contiguous segments consisting of 4 timesteps, which are then randomized and combined along the batch dimension.
The first observation in each segment is the initial condition for the prediction.

With the exception of the power grid experiments, all trajectories are generated using a 9(8) explicit Runge--Kutta method,\cite{verner_numerically_2010} with absolute and relative solver tolerances of $10^{-12}$.

The power grid models, which are differential-algebraic equations (DAEs), are implemented in the Julia package \texttt{PowerDynamics.jl}.\cite{plietzsch_powerdynamicsjlexperimentally_2022}
Given a random initial condition, we then generate a ground truth trajectory by solving the DAE using a 4th order L-stable Rosenbrock method.

\subsection{Initial Conditions}

\paragraph{Quartic Potential.}
Initial positions are drawn uniformly from a ring $r \in [0.8, 1.2]$ with random angle, and initial velocities are set to produce circular orbits with small random perturbations.

\hfill

\paragraph{Lotka--Volterra.}
Initial populations are drawn from $x_0, y_0 \sim \mathrm{Unif}(0.5, 2.0)$.

\hfill

\paragraph{$N$-Pendulum.}
The initial displacements of the pendulum arms are drawn independently from $\mathrm{Unif}(0,2\pi)$, and the pendulums are released from rest (zero initial velocity).

\hfill

\paragraph{Power Grids.}
Starting from a stable operation point, we generate perturbed initial conditions using the ambient forcing method,\cite{buttner_ambient_2022} which preserves the algebraic PQ-bus constraints by construction.
Following the approach of \citet{nauck_predicting_2026}, each bus~$j$ is assigned a random forcing vector $(F_j^{\mathrm{re}}, F_j^{\mathrm{im}}) = r_j(\cos\theta_j, \sin\theta_j)$, where $\theta_j \sim \mathrm{Unif}(0,2\pi)$ and $r_j^2 \sim \mathrm{Unif}(a, b)$ with $a = 0.15$, $b = 1$.
Sampling $r_j^2$ (rather than $r_j$) uniformly ensures that the forcing vectors are distributed with uniform density per unit area in the complex voltage plane, so that large-magnitude perturbations are not under-represented relative to their share of the annulus area.
The composite forcing vector~$F$ is applied for a short interval~$\tau = 0.2$, and the subsequent unforced transient dynamics provide the training trajectories.

\subsection{State and Constraint Dimensions}
Table~\ref{tab:dimensions} summarizes the state dimension~$n$ and number of constraints~$m$ for each system.
For the $N$-pendulum systems, dimensions correspond to the Cartesian coordinate formulation used by SNDE and PNDE.
\begin{table}[h]
  \caption{State dimension~$n$ and number of constraints~$m$ for each system.}
  \label{tab:dimensions}
  \begin{ruledtabular}
    \begin{tabular}{lcc}\noalign{\vspace{0.5ex}}
      System              & Dimension $n$ & Constraints $m$ \\[0.5ex]
      \hline\noalign{\vspace{0.5ex}}
      2-D Quartic Potential & 4  & 1  \\
      Lotka--Volterra        & 2  & 1  \\
      $4$-Pendulum          & 16 & 8  \\
      $8$-Pendulum          & 32 & 16 \\
      IEEE 14-Bus           & 28 & 18 \\
    \end{tabular}
  \end{ruledtabular}
\end{table}
\subsection{Normalization}
Normalization of the data requires careful attention.
For all datasets, we standardize the data to have zero mean and unit standard deviation.
However, to evaluate and apply the constraints, we require the state in physical units.
Accordingly, we solve the NDE in physical units, meaning that we also evaluate and apply the constraints in physical units.
This requires de-normalizing the initial condition before solving and then re-normalizing the predicted trajectory before computing the loss (these operations are cheap).
Despite solving in physical units, the inputs to the MLP are always normalized; the MLP takes a normalized state and predicts velocities in physical units.
This requires continually normalizing the (physical) state prediction from the previous solver step before providing it as input to the MLP in the next step.

The only exception is the $N$-pendulum NDE in generalized coordinates.
Here, we provide as MLP inputs the sines and cosines of the angles, rather than the angles themselves.
Since the sines and cosines are bounded between -1 and 1, we do not normalize these inputs, but we do normalize the angular velocities as usual.
\section{Hyperparameters and Training}
\label{app:training}
We summarize the hyperparameters for each system in Table~\ref{tab:data-hyperpar}.
We minimize the mean squared error loss using the AdamW optimizer.\cite{loshchilov_decoupled_2019}
All systems are modeled using standard multilayer perceptrons with GELU activation functions to encourage smoothness.
For the $N$-pendulum experiments, we also include skip connections between the hidden layers to improve training stability.
The MLP hyperparameters in Table~\ref{tab:data-hyperpar} were selected by a grid search over the unconstrained NDE for each system, and the same settings were then reused for the corresponding constrained models (SNDE and PNDE) to ensure a fair comparison.

During training, all models are solved using a 5(4) explicit Runge--Kutta method\cite{tsitouras_rungekutta_2011} with absolute and relative tolerances of $10^{-4}$.
Gradients of ODE solutions are obtained using the adjoint sensitivity method.
For testing, we use absolute and relative tolerances of $10^{-6}$.

\begin{table}
  \caption{Summary of datasets and hyperparameters.}
  \label{tab:data-hyperpar}
  \begin{ruledtabular}
    \begin{tabular}{ccccc}\noalign{\vspace{0.5ex}}
      & \shortstack{2-D Quartic                                            \\Potential} & \shortstack{Lotka-\\Volterra} & \raisebox{1.3ex}{$N$-pendulum} & \shortstack{IEEE\\14-Bus} \\[0.5ex]
      \hline\noalign{\vspace{0.5ex}}
      $n_\mathrm{train}$ & 120                     & 120       & 400              & 100       \\
      $n_\mathrm{valid}$ & 60                      & 60        & 200              & 50        \\
      $n_\mathrm{test}$  & 100                     & 100       & 100              & 100       \\
      $T$                & 25                      & 25        & 10               & 10        \\
      $dt$               & 0.1                     & 0.1       & 0.01             & 0.1       \\[0.5ex]
      \hline\noalign{\vspace{0.5ex}}
      Hidden Layers      & 4                       & 6         & 10               & 10        \\
      Hidden Width       & 512                     & 512       & 1024             & 512       \\
      Learning Rate      & $10^{-3}$               & $10^{-3}$ & $5\times10^{-4}$ & $10^{-3}$ \\
      Weight Decay       & $10^{-4}$               & $10^{-4}$ & $10^{-4}$        & $10^{-4}$ \\
      Batch Size         & 1024                    & 1024      & 1024             & 1024      \\
      Epochs             & 1000                    & 1000      & 1000             & 1000      \\
    \end{tabular}
  \end{ruledtabular}
\end{table}

\section{Supplementary Figures}
\label{app:supplementary_figures}
\begin{figure}
  \centering
  \includegraphics[width=\columnwidth]{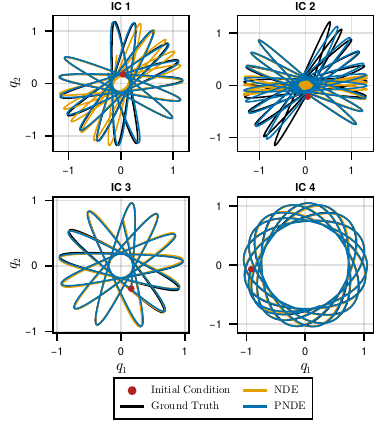}
  \caption{
    Trajectories of the 2-D quartic potential system, comparing ground truth with NDE and PNDE predictions for the first four initial conditions in the test set.
  }
  \label{fig:quartic_potential_phase}
\end{figure}

\begin{figure}
  \centering
  \includegraphics[width=\columnwidth]{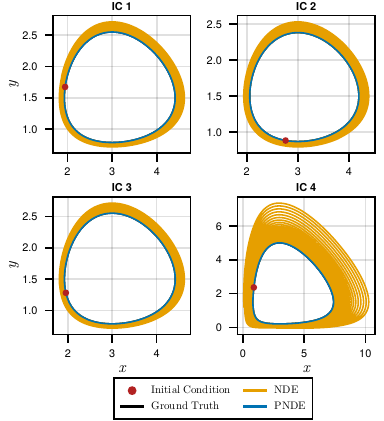}
  \caption{
    Trajectories of the Lotka--Volterra system, comparing ground truth with NDE and PNDE predictions for the first four initial conditions in the test set.
  }
  \label{fig:lotka_volterra_phase}
\end{figure}

\begin{figure}
  \centering
  \includegraphics[width=\columnwidth]{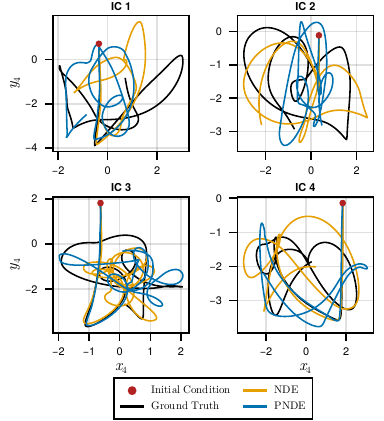}
  \caption{
    Trajectories of the 4-pendulum system, comparing ground truth with NDE and PNDE predictions for the first four initial conditions in the test set.
    We plot the position of the last pendulum bob as the pendulum is released from rest.
  }
  \label{fig:pendulum_phase_4}
\end{figure}

\begin{figure}
  \centering
  \includegraphics[width=\columnwidth]{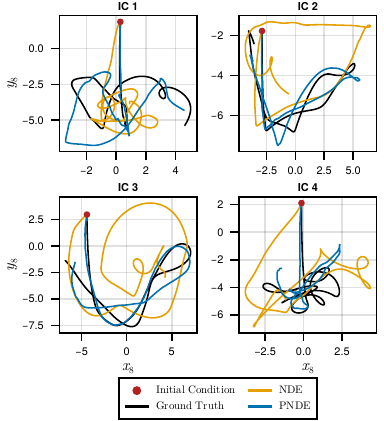}
  \caption{
    Trajectories of the 8-pendulum system, comparing ground truth with NDE and PNDE predictions for the first four initial conditions in the test set.
    We plot the position of the last pendulum bob as the pendulum is released from rest.
  }
  \label{fig:pendulum_phase_8}
\end{figure}

For each system, we plot in Figs.~\ref{fig:quartic_potential_phase}--\ref{fig:pendulum_phase_8} the trajectories of the first four initial conditions in the test set, comparing the ground truth with NDE and PNDE predictions.
In all cases, the PNDE produces visibly better predictions than the NDE.
\begin{figure}[t]
  \centering
  \includegraphics[width=\columnwidth]{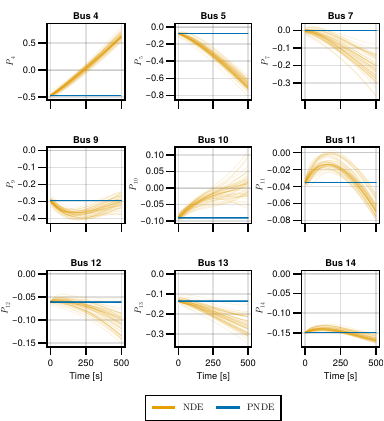}
  \caption{
    Active power injection at all nine PQ buses of the IEEE 14-bus system over $500\,\mathrm{s}$ rollouts.
    At every bus, the NDE trajectories (orange) drift away from their initial power injection values, while the PNDE trajectories (blue) remain constant, reflecting the constraints.
  }
  \label{fig:ieee14bus-long-rollout-supp-pq}
\end{figure}
\begin{figure}[t]
  \centering
  \includegraphics[width=\columnwidth]{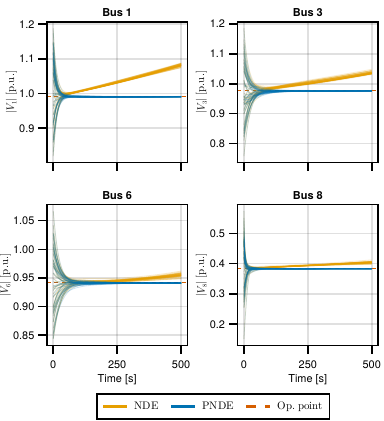}
  \caption{
    Voltage magnitude at the four PV buses of the IEEE 14-bus system over $500\,\mathrm{s}$ rollouts.
    The dashed line indicates the operation point.
    PNDE trajectories (blue) converge to the operation point after initial transients, while NDE trajectories (orange) diverge.
  }
  \label{fig:ieee14bus-long-rollout-supp-pv}
\end{figure}
\section{Constraint Jacobians}
\label{appendix:jacobians}
We provide here the analytic constraint Jacobian~$G(u)$ for each system.
All Jacobians were tested for correctness using automatic differentiation.

\subsection{2-D Quartic Potential}

For the energy constraint $g(u) = E(u) - E(u_0)$ with $u = (x, y, \dot{x}, \dot{y})$ and $E$ as in Eq.~\eqref{eq:quartic_energy}, the Jacobian $G \in \sR^{1 \times 4}$ is
\begin{equation}
  G =
  \begin{pmatrix} x(x^2{+}y^2) & y(x^2{+}y^2) & \dot{x} & \dot{y}
  \end{pmatrix}.
\end{equation}

\subsection{Lotka--Volterra}

For the conserved quantity constraint $g(u) = V(u) - V(u_0)$ with $u = (x,y)$ and $V$ as in Eq.~\eqref{eq:lv_conserved}, the Jacobian $G \in \sR^{1 \times 2}$ is
\begin{equation}
  G =
  \begin{pmatrix} \delta - \gamma/x & \beta - \alpha/y
  \end{pmatrix}.
\end{equation}

\subsection{\texorpdfstring{$N$}{N}-Pendulum}

With state $u = (x, y, \dot{x}, \dot{y}) \in \sR^{4N}$ in Cartesian coordinates, where $x = (x_1,\ldots,x_N)^\top$ collects the horizontal positions and similarly for $y$, $\dot{x}$, $\dot{y}$, define the relative displacements
\begin{equation}
  \begin{aligned}
    \Delta x_i       & = x_i - x_{i-1}, \quad             & \Delta y_i       & = y_i - y_{i-1},             \\
    \Delta \dot{x}_i & = \dot{x}_i - \dot{x}_{i-1}, \quad & \Delta \dot{y}_i & = \dot{y}_i - \dot{y}_{i-1},
  \end{aligned}
\end{equation}
with $x_0 = y_0 = \dot{x}_0 = \dot{y}_0 = 0$.
The Jacobian $G \in \sR^{2N \times 4N}$ of the position and velocity constraints~\eqref{eq:holonomic}--\eqref{eq:non_holonomic} has the block structure
\begin{equation}
  \label{eq:npend_jac}
  G =
  \begin{pmatrix}
    2\, D_{\Delta x}   & 2\, D_{\Delta y}   & 0            & 0            \\
    D_{\Delta \dot{x}} & D_{\Delta \dot{y}} & D_{\Delta x} & D_{\Delta y}
  \end{pmatrix},
\end{equation}
where $D_a \coloneqq \mathrm{diag}(a)\, D$ for any $a \in \sR^N$, $\mathrm{diag}(a)$ denotes the diagonal matrix formed from~$a$, and $D \in \sR^{N \times N}$ is the lower bidiagonal first-difference matrix
\begin{equation}
  D_{ij} = \delta_{ij} - \delta_{i-1,\,j}.
\end{equation}
Each $N \times N$ block in Eq.~\eqref{eq:npend_jac} is bidiagonal, reflecting the chain topology of the pendulum.

\subsection{Power Grids}

Writing the state as $u = (v^{re}, v^{im}) \in \sR^{2N}$, where $v^{re}, v^{im} \in \sR^N$ collect the real and imaginary parts of the complex bus voltages, the complex current is $i = \mathrm{LY}\, v$ with real and imaginary parts
\begin{equation}
  i^{re} = Y^{re}\, v^{re} - Y^{im}\, v^{im}, \quad i^{im} = Y^{re}\, v^{im} + Y^{im}\, v^{re},
\end{equation}
where $Y^{re} = \Real(\mathrm{LY})$ and $Y^{im} = \Imag(\mathrm{LY})$.
Given $M$ PQ-buses with index set~$\mathcal{P}$, the active and reactive power at bus~$j$ are $P_j = v_j^{re}\, i_j^{re} + v_j^{im}\, i_j^{im}$ and $Q_j = v_j^{im}\, i_j^{re} - v_j^{re}\, i_j^{im}$.
Up to an overall sign arising from the constraint definition $g = (P_j - \ldots, Q_j - \ldots)$, which leaves the projection operator in Eq.~(\ref{eq:proj_matrix}) unchanged, the Jacobian $G \in \sR^{2M \times 2N}$ has entries, for $j \in \mathcal{P}$ and $k = 1,\ldots, N$,
\begin{subequations}
  \label{eq:power_jac}
  \begin{align}
    \frac{\partial P_j}{\partial v_k^{re}} & = \delta_{jk}\, i_j^{re} + v_j^{re}\, Y_{jk}^{re} + v_j^{im}\, Y_{jk}^{im},  \\
    \frac{\partial P_j}{\partial v_k^{im}} & = \delta_{jk}\, i_j^{im} - v_j^{re}\, Y_{jk}^{im} + v_j^{im}\, Y_{jk}^{re},  \\
    \frac{\partial Q_j}{\partial v_k^{re}} & = -\delta_{jk}\, i_j^{im} - v_j^{re}\, Y_{jk}^{im} + v_j^{im}\, Y_{jk}^{re}, \\
    \frac{\partial Q_j}{\partial v_k^{im}} & = \delta_{jk}\, i_j^{re} - v_j^{re}\, Y_{jk}^{re} - v_j^{im}\, Y_{jk}^{im}.
  \end{align}
\end{subequations}
\bibliography{pnde}
\end{document}